# Digital Color Imaging

Gaurav Sharma, *Member, IEEE,* and H. Joel Trussell, *Fellow, IEEE*

*Abstract*—This paper surveys current technology and research in the area of digital color imaging. In order to establish the background and lay down terminology, fundamental concepts of color perception and measurement are first presented using vector-space notation and terminology. Present-day color recording and reproduction systems are reviewed along with the common mathematical models used for representing these devices. Algorithms for processing color images for display and communication are surveyed, and a forecast of research trends is attempted. An extensive bibliography is provided.

## I. Introduction

AMONG THE human senses, sight and color perception are perhaps the most fascinating. There is, consequently, little wonder that color images pervade our daily life in television, photography, movies, books, and newspapers. With the digital revolution, color has become even more accessible. Color scanners, cathode ray tube (CRT) displays, and printers are now an integral part of the office environment. Extrapolating from current trends, homes will also have a plethora of digital color imaging products in the near future.

The increased use of color has brought with it new challenges and problems. In order to meaningfully record and process color images, it is essential to understand the mechanisms of color vision and the capabilities and limitations of color imaging devices. It is also necessary to develop algorithms that minimize the impact of device limitations and preserve color information as images are exchanged between devices. The goal of this paper is to present a survey of the technology and research in these areas.

The rest of this paper is broadly organized into four sections. Section II provides an introduction to color science for imaging applications. Commonly used color recording and reproduction devices are discussed in Section III. A survey of algorithms used for processing color images in desktop applications is presented in Section IV. Finally, research directions in color imaging are summarized in Section V.

## II. Color Fundamentals

Prior to the time of Sir Isaac Newton, the nature of light and color was rather poorly understood [1], [2]. Newton's meticulous experiments [3], [4, Chap. 3] with sunlight and a prism helped dispel existing misconceptions and led to the realization that the color of light depended on its spectral composition. Even though Grimaldi preceded Newton in making these discoveries, his book [5], [2, pp. 141–147] on the subject received attention much later, and credit for the widespread dissemination of the new ideas goes to Newton. While Newton's experiments established a physical basis for color, they were still a long way from a system for colorimetry.

Before a system to measure and specify color could be developed, it was necessary to understand the nature of the color sensing mechanisms in the human eye. While some progress in this direction was made in the late 18th century [6], the prevalent anthropocentric views contributed to a confusion between color vision and the nature of light [6], [7]. The wider acceptance of the wave theory of light paved the way for a better understanding of both light and color [8], [9]. Both Palmer [6] and Young [9] hypothesized that the human eye has three receptors, and the difference in their responses contributes to the sensation of color. However, Grassmann [10] and Maxwell [11] were the first to clearly state that color can be mathematically specified in terms of three independent variables. Grassmann also stated experimental laws of color matching that now bear his name [12, p. 118]. Maxwell [13], [14] demonstrated that any additive color mixture could be "matched" by proper amounts of three primary stimuli, a fact now referred to as *trichromatic generalization* or *trichromacy*. Around the same time, Helmholtz [15] explained the distinction between additive and subtractive color mixing and explained trichromacy in terms of spectral sensitivity curves of the three "color sensing fibers" in the eye.

Trichromacy provided strong indirect evidence for the fact that the human eye has three color receptors. This fact was confirmed only much later by anatomical and physiological studies. The three receptors are known as the S, M, and L cones (*short, medium,* and *long* wavelength sensitive) and their spectral sensitivities have now been determined directly through microspectrophotometric measurements [16], [17]. Long before these measurements were possible, *color-matching functions* (CMF's) were determined through psychophysical experiments [12], [18]–[21]. CMF's are sets of three functions related to the spectral sensitivities of the three cones by nonsingular linear transformations. The CMF's determined by Guild [19] and Wright [18] were used by the CIE (Commission Internationale de l'Éclairage or the International Commission on Illumination) to establish a standard for a numerical specification of color in terms of three coordinates or *tristimulus values*.

While the CMF's provide a basis for a linear model for color specification, it is clear that the human visual sensitivity to color changes is nonlinear. Since color differences between real world objects and images are evaluated by human viewers, it is desirable to determine *uniform color spaces* in which equal Euclidean distances correspond to roughly equal perceived color differences. Considerable research has focused on this problem since the establishment of colorimetry.

Tristimulus values are useful for specifying colors and communicating color information precisely. Uniform color spaces are useful in evaluating color matching/mismatching of similar stimuli under identical adaptation conditions. Since the human visual system undergoes significant changes in response to its environment, tristimuli under different conditions of adaptation cannot be meaningfully compared. Since typical color reproduction problems involve different media or viewing conditions, it is necessary to consider descriptors of *color appearance* that transcend these adaptations. This is the goal of color appearance modeling.

### A. Trichromacy and Human Color Vision

In the human eye, an image is formed by light focused onto the retina by the eye's lens. The three types of cones that govern color sensation are embedded in the retina, and contain photosensitive pigments with different spectral absorptances. If the spectral distribution of light incident on the retina is given by $f(\lambda)$, where $\lambda$ represents wavelength (we are ignoring any spatial variations in the light for the time being), the responses of the three cones can be modeled as a three vector with components given by

$$c_i = \int_{\lambda_{\min}}^{\lambda_{\max}} s_i(\lambda) f(\lambda) \, d\lambda \quad i = 1, 2, 3 \qquad (1)$$

where $s_i(\lambda)$ denotes the sensitivity of the $i$th type of cones, and $[\lambda_{\min}, \lambda_{\max}]$ denotes the interval of wavelengths outside of which all these sensitivities are zero. Typically in air or vacuum, the visible region of the electromagnetic spectrum is specified by the wavelength region between $\lambda_{\min} = 360$ nm and $\lambda_{\max} = 830$ nm.

Mathematically, the expressions in (1) correspond to inner product operations [22] in the Hilbert space of square integrable functions $\mathcal{L}^2([\lambda_{\min}, \lambda_{\max}])$. Hence, the cone response mechanism corresponds to a projection of the spectrum onto the space spanned by three sensitivity functions $\{s_i(\lambda)\}_{i=1}^{3}$. This space is called the *human visual subspace* (HVSS) [23]–[26]. The perception of color depends on further nonlinear processing of the retinal responses. However, to a first order of approximation, under similar conditions of adaptation, the sensation of color may be specified by the responses of the cones. This is the basis of all colorimetry and will be implicitly assumed throughout this section. A discussion of perceptual uniformity and appearance will be postponed until Sections II-C and II-D.

For computation, the spectral quantities in (1) may be replaced by their sampled counterparts to obtain summations as numerical approximations to the integrals. For most color spectra, a sampling rate of 10 nm provides sufficient accuracy, but in applications involving fluorescent lamps with sharp spectral peaks, a higher sampling rate or alternative approaches may be required [27]–[30].

If $N$ uniformly spaced samples are used over the visible range $[\lambda_{\min}, \lambda_{\max}]$, (1) can be compactly written as

$$\mathbf{c} = \mathbf{S}^T \mathbf{f} \qquad (2)$$

where the superscript $T$ denotes the transpose, $\mathbf{c} = [c_1, c_2, c_3]^T$, $\mathbf{S}$ is an $N \times 3$ matrix whose $i$th column, $\mathbf{s}_i$, is the vector of samples of $s_i(\lambda)$, and $\mathbf{f}$ is the $N \times 1$ vector of samples of $f(\lambda)$. The HVSS then corresponds to the column space of $\mathbf{S}$.

In normal human observers, the spectral sensitivities of the three cones are linearly independent. Furthermore, the differences between the spectral sensitivities of color-normal observers are (relatively) small [18], [31], [12, p. 343] and arise primarily due to the difference in the spectral transmittance of the eye's lens and the optical medium ahead of the retina [18], [32]–[34].

If a standardized set of cone responses is defined, color may be specified using the three-vector, $\mathbf{c}$, in (2), known as a *tristimulus* vector. Just as several different coordinate systems may be used for specifying position in three-dimensional (3-D) space, any nonsingular well-defined linear transformation of the tristimulus vector, $\mathbf{c}$, can also serve the purpose of color specification. Since the cone responses are difficult to measure directly, but nonsingular linear transformations of the cone responses are readily determined through color-matching experiments, such a transformed coordinate system is used for the measurement and specification of color.

*1) Color Matching:* Two spectra, represented by $N$-vectors, $\mathbf{f}$ and $\mathbf{g}$, produce the same cone responses and therefore represent the same color if

$$\mathbf{S}^T \mathbf{f} = \mathbf{S}^T \mathbf{g}. \qquad (3)$$

To see how (2) encapsulates the principle of trichromacy and how CMF's are determined, consider three color *primaries*, i.e., three *colorimetrically independent* light sources $\mathbf{p}_1, \mathbf{p}_2, \mathbf{p}_3$. The term *colorimetrically independent* will be used in this paper to denote a collection of spectra such that the color of any one cannot be visually matched by any linear combination of the others. Mathematically, colorimetric independence of $\mathbf{p}_1, \mathbf{p}_2, \mathbf{p}_3$ is equivalent to the linear independence of the three-vectors $\mathbf{S}^T \mathbf{p}_1, \mathbf{S}^T \mathbf{p}_2$, and $\mathbf{S}^T \mathbf{p}_3$. Hence if $\mathbf{P} = [\mathbf{p}_1, \mathbf{p}_2, \mathbf{p}_3]$, the $3 \times 3$ matrix $\mathbf{S}^T \mathbf{P}$ is nonsingular.

For any visible spectrum, $\mathbf{f}$, the three-vector $\mathbf{a}(\mathbf{f}) \overset{\text{def}}{=} (\mathbf{S}^T \mathbf{P})^{-1} \mathbf{S}^T \mathbf{f}$ satisfies the relation

$$\mathbf{S}^T \mathbf{f} = \mathbf{S}^T \mathbf{P} \mathbf{a}(\mathbf{f}) \qquad (4)$$

which is the relation for a color match. Hence, for any visible spectrum, $\mathbf{f}$, there exists a combination of the primaries, $\mathbf{P}\mathbf{a}(\mathbf{f})$, which matches the color of $\mathbf{f}$. This statement encapsulates the principle of trichromacy. From the standpoint of obtaining a physical match, the above mathematical argument requires some elaboration. It is possible that the obtained vector of primary "strengths," $\mathbf{a}(\mathbf{f})$, has negative components (in fact it can be readily shown that for any set of physical primaries there exist visible spectra for which this happens).

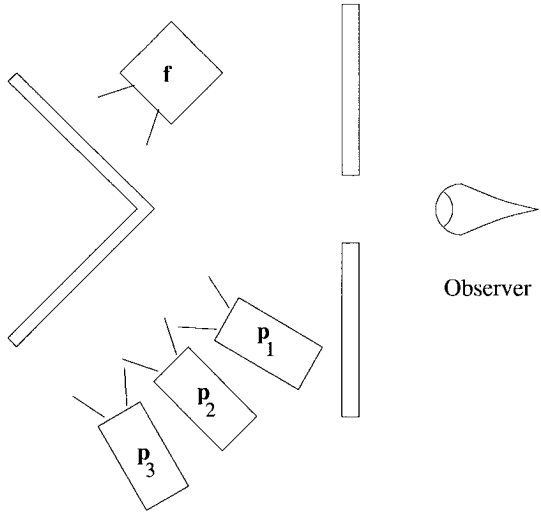

Fig. 1. Color-matching experiment.

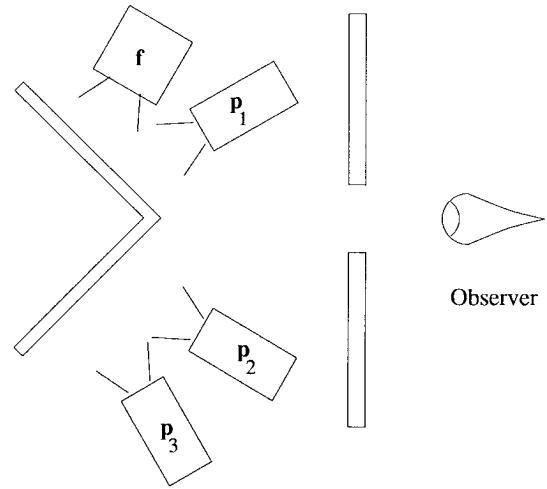

Fig. 2. Color-matching experiment with negative value for primary $\mathbf{p}_1$.

Since negative intensities of the primaries cannot be produced, the spectrum $\mathbf{Pa}(\mathbf{f})$ is not realizable using the primaries. A physical realization corresponding to the equations is, however, still possible by rearranging the terms in (4) and "subtracting" the primaries with negative "strengths" from $\mathbf{f}$. The double negation cancels out and corresponds to the addition of positive amounts of the appropriate primaries to $\mathbf{f}$.

The setup for a typical color-matching experiment is shown schematically in Fig. 1. The observer views a small circular field that is split into two halves. The spectrum $\mathbf{f}$ is displayed on one half of a visual field. On the other half of the visual field appears a linear combination of the primary sources. The observer attempts to visually match the input spectrum by adjusting the relative intensities of the primary sources. The vector, $\mathbf{a}(\mathbf{f})$, denotes the relative intensities of the three primaries when a match is obtained. Physically, it may be impossible to match the input spectrum by adjusting the intensities of the primaries. When this happens, the observer is allowed to move one or two of the primaries so that they illuminate the same field as input spectrum, $\mathbf{f}$ (see Fig. 2). As noted earlier, this procedure is mathematically equivalent to subtracting that amount of primary from the primary field, i.e., the strengths in $\mathbf{a}(\mathbf{f})$ corresponding to the primaries which were moved are negative. As demonstrated in the last paragraph, all visible spectra can be matched using this method.

*2) Color-Matching Functions:* The linearity of color matching expressed in (3) implies that if the color tristimulus values for a basis set of spectra are known, the color values for all linear combinations of those spectra can be readily deduced. The unit intensity monochromatic spectra, given by $\{\mathbf{e}_i\}_{i=1}^{N}$, where $\mathbf{e}_i$ is an $N$-vector having a one in the $i$th position and zeros elsewhere, form an orthonormal basis in terms of which all spectra can be expressed. Hence, the color matching properties of all spectra (with respect to a given set of primaries) can be specified in terms of the color matching properties of these monochromatic spectra.

Consider the color-matching experiment of the last section for the monochromatic spectra. Denoting the relative intensities of the three primaries required for matching $\mathbf{e}_i$ by $\mathbf{a}_i = \mathbf{a}(\mathbf{e}_i)$, the matches for all the monochromatic spectra can be written as

$$\mathbf{S}^T\mathbf{e}_i = \mathbf{S}^T\mathbf{Pa}_i \quad i = 1, 2, \dots N. \tag{5}$$

Combining the results of all $N$ monochromatic spectra, we get

$$\mathbf{S}^T\mathbf{I} = \mathbf{S}^T\mathbf{PA}^T \tag{6}$$

where $\mathbf{I} = [\mathbf{e}_1, \mathbf{e}_2, \dots, \mathbf{e}_N]$ is the $N \times N$ identity matrix, and $\mathbf{A} = [\mathbf{a}_1, \mathbf{a}_2, \dots, \mathbf{a}_N]^T$ is the *color-matching matrix* corresponding to the primaries $\mathbf{P}$. The entries in the $k$th column of $\mathbf{A}$ correspond to the relative amount of the $k$th primary required to match $\{\mathbf{e}_i\}_{i=1}^{N}$, respectively. The columns of $\mathbf{A}$ are therefore referred to as the *color-matching functions* (CMF's) (associated with the primaries $\mathbf{P}$).

From (6), it can be readily seen that the color-matching matrix $\mathbf{A} = \mathbf{S}(\mathbf{P}^T\mathbf{S})^{-1}$. Hence the CMF's are a nonsingular linear transformation of the sensitivities of the three cones in the eye. It also follows that the color of two spectra, $\mathbf{f}$ and $\mathbf{g}$, matches if and only if $\mathbf{A}^T\mathbf{f} = \mathbf{A}^T\mathbf{g}$. As mentioned earlier, color of a visible spectrum, $\mathbf{f}$, may be specified in terms of the tristimulus values, $\mathbf{A}^T\mathbf{f}$, instead of $\mathbf{S}^T\mathbf{f}$. The fact that the color-matching matrix is readily determinable using the procedure outlined above makes such a scheme for specifying color considerably attractive in comparison to one based on the actual cone sensitivities. Note also that the HVSS that was defined as the column space of $\mathbf{S}$ can alternately be defined as the column space of $\mathbf{A}$.

*3) Metamerism and Black Space:* As stated in (3), two spectra represented by $N$-vectors $\mathbf{f}$ and $\mathbf{g}$ match in color if $\mathbf{S}^T\mathbf{f} = \mathbf{S}^T\mathbf{g}$ (or $\mathbf{A}^T\mathbf{f} = \mathbf{A}^T\mathbf{g}$). Since $\mathbf{S}$ (or equivalently $\mathbf{A}$) is an $N \times 3$ matrix, with $N > 3$, it is clear that there are several different spectra that appear to be the same color to the observer. Two distinct spectra that appear the same are called *metamers*, and such a color match is said to be a *metameric match* (as opposed to a spectral match).

Metamerism is both a boon and a curse in color applications. Most color output systems (such as CRT's and color photography) exploit metamerism to reproduce color. However, in the matching of reflective materials, a metameric match under one viewing illuminant is usually insufficient to establish a match under other viewing illuminants. A common manifestation of this phenomenon is the color match of (different) fabrics under one illumination and mismatch under another.

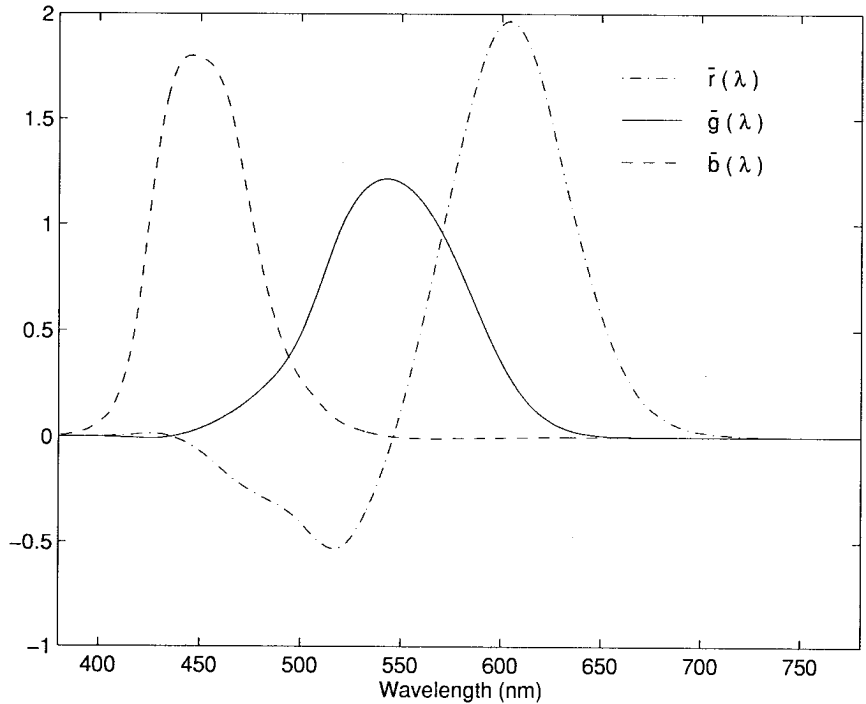

Fig. 3.   CIE $\bar{r}(\lambda)$, $\bar{g}(\lambda)$, $\bar{b}(\lambda)$ color-matching functions.

The vector space view of color matching presented above was first given by Cohen and Kaupauf [35], [36], [24]. Tutorial descriptions using current notation and terminology appear in [23], [25], [37], and [38]. This approach allows us to deduce a number of interesting and useful properties of color vision. One such property is the decomposition of the $N$ dimensional spectral space into the 3-D HVSS and the $(N-3)$-dimensional *metameric black space*, which was first hypothesized by Wyszecki [39]. Mathematically, this result states that any visible spectrum, $\mathbf{f}$, can be written as

$$\mathbf{f} = \mathbf{P_A f} + \mathbf{P_A^\perp f} \tag{7}$$

where $\mathbf{P_A} = \mathbf{A}(\mathbf{A}^T \mathbf{A})^{-1}\mathbf{A}^T$ is the orthogonal projector onto the column space of $\mathbf{A}$, i.e., the HVSS, and $\mathbf{P_A^\perp} = (\mathbf{I} - \mathbf{P_A})$ is the orthogonal projector onto the black space, which is the orthogonal complement of the HVSS. The projection, $\mathbf{P_A f}$, is called the *fundamental metamer* of $\mathbf{f}$ because all metamers of $\mathbf{f}$ are given by $\{\mathbf{P_A f} + \mathbf{P_A^\perp g} \mid \mathbf{g} \in R^N\}$.

Another direct consequence of the above description of color matching is the fact that the primaries in any color-matching experiment are unique only up to metamers. Since metamers are visually identical, the CMF's are not changed if each of the three primaries are replaced by any of their metamers.

The physical realization of metamers imposes additional constraints over and above those predicated by the equations above. In particular, any physically realizable spectrum needs to be nonnegative, and, hence, it is possible that the metamers described by the above mathematics may not be realizable. In cases where a realizable metamer exists, set theoretic approaches may be used to incorporate nonnegativity and other constraints [37].

### B. Colorimetry

It was mentioned in Section II-A2 that the color of a visible spectrum, $\mathbf{f}$, can be specified in terms of the tristimulus values, $\mathbf{A}^T \mathbf{f}$, where $\mathbf{A}$ is a matrix of CMF's. In order to have agreement between different measurements, it is necessary to define a standard set of CMF's with respect to which the tristimulus values are stated. A number of different standards have been defined for a variety of applications, and it is worth reviewing some of these standards and the historical reasons behind their development.

*1) CIE Standards:* The CIE is the primary organization responsible for standardization of color metrics and terminology. A colorimetry standard was first defined by the CIE in 1931 and continues to form the basis of modern colorimetry.

The CIE 1931 recommendations define a standard colorimetric observer by providing two different but equivalent sets of CMF's. The first set of CMF's is known as the *CIE Red–Green–Blue (RGB)* CMF's, $\bar{r}(\lambda)$, $\bar{g}(\lambda)$, $\bar{b}(\lambda)$. These are associated with monochromatic primaries at wavelengths of 700.0, 546.1, and 435.8 nm, respectively, with their radiant intensities adjusted so that the tristimulus values of the equi-energy spectrum are all equal [40]. The equi-energy spectrum is the one whose spectral irradiance (as a function of wavelength) is constant. The CIE RGB CMF's are shown in Fig. 3.

The second set of CMF's, known as the *CIE XYZ* CMF's, are $\bar{x}(\lambda)$, $\bar{y}(\lambda)$, and $\bar{z}(\lambda)$; they are shown in Fig. 4. They were recommended for reasons of more convenient application in colorimetry and are defined in terms of a linear transformation of the CIE RGB CMF's [41]. When these CMF's were first defined, calculations were typically performed on desk calculators, and the repetitive summing and differencing due to the negative lobes of the CIE RGB CMF's was prone to errors. Hence, the transformation from the CIE RGB CMF's to CIE XYZ CMF's was determined so as to avoid negative values at all wavelengths [42]. Since an infinite number of transformations can be defined in order to meet this non-negativity requirement, additional criteria were used in the choice of the CMF's [43], [44, p. 531]. Two of the important considerations were the choice of $\bar{y}(\lambda)$ coincident with the *luminous efficiency function* [12] and the normalization of the three CMF's so as to yield equal tristimulus values for the equi-energy spectrum. The luminous efficiency function gives the relative sensitivity of the eye to the energy at each wavelength. From the discussion of Section II-A1, it is readily seen that CMF's that are nonnegative for all wavelengths cannot be obtained with any physically realizable primaries. Hence, any set of primaries corresponding to the CIE XYZ CMF's is not physically realizable.

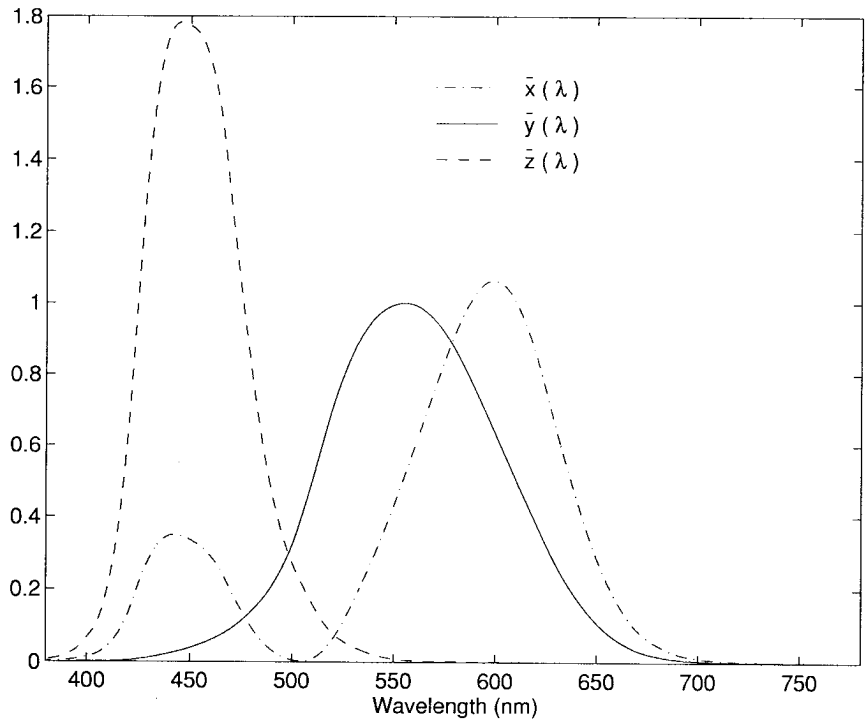

Fig. 4. CIE $\bar{x}(\lambda)$, $\bar{y}(\lambda)$, $\bar{z}(\lambda)$ color matching functions.

The tristimulus values obtained with the CIE RGB CMF's are called the *CIE RGB tristimulus values*, and those obtained with the CIE XYZ CMF's are called the *CIE XYZ tristimulus values*. The $Y$ tristimulus value is usually called the *luminance* and correlates with the perceived brightness of the radiant spectrum.

The two sets of CMF's described above are suitable for describing color matching when the angular subtense of the matching fields at the eye is between one and four degrees [12, p. 131], [40, p. 6]. When the inadequacy of these CMF's for matching fields with larger angular subtense became ap-parent, the CIE defined an alternate standard colorimetric observer in 1964 with different sets of CMF's [40]. Since imaging applications (unlike quality control applications in manufacturing) involve complex visual fields where the color-homogeneous areas have small angular subtense, the CIE 1964 (10° observer) CMF's will not be discussed here.

In addition to the CMF's, the CIE has defined a number of standard illuminants for use in colorimetry of nonluminous reflecting objects. The relative irradiance spectra of a number of these standard illuminants is shown in Fig. 5. To represent different phases of daylight, a continuum of daylight illuminants has been defined [40], which are uniquely specified in terms of their *correlated color temperature*. The correlated color temperature of an illuminant is defined as the temperature of a *black body radiator* whose color most closely resembles that of the illuminant [12]. D65 and D50 are two daylight illuminants commonly used in colorimetry, which correspond to correlated color temperatures of 6500 and 5000 K, respectively. The CIE illuminant A represents a black body radiator at a temperature of 2856 K and closely approximates the spectra of incandescent lamps.

A nonluminous object is represented by the $N$-vector, $\mathbf{r}$, of samples of its spectral reflectance, where $0 \leq r_i \leq 1, 1 \leq i \leq N$. When the object is viewed under an illuminant with spectrum given by the $N$ vector, $\mathbf{l}$, the resulting spectral radiance at the eye is obtained as the product of the illuminant spectrum and the reflectance at each wavelength. Therefore, the CIE XYZ tristimulus values defining the color are given by

$$\mathbf{t} = \mathbf{A}^T \mathbf{L} \mathbf{r} = \mathbf{A}_\mathbf{L}^T \mathbf{r} \qquad (8)$$

where $\mathbf{A}$ is the matrix of CIE XYZ CMF's, $\mathbf{L}$ is the diagonal illuminant matrix with entries from $\mathbf{1}$ along the diagonal, and $\mathbf{A_L = LA}$. In analogy with the HVSS, the column space of $\mathbf{A_L}$ is defined as the *human visual illuminant subspace* (HVISS) [26]. Note that (8) is based on an idealized model of illuminant–object interaction that does not account for several geometry/surface effects such as the combination of specular and body reflectance components [30, pp. 43–45].

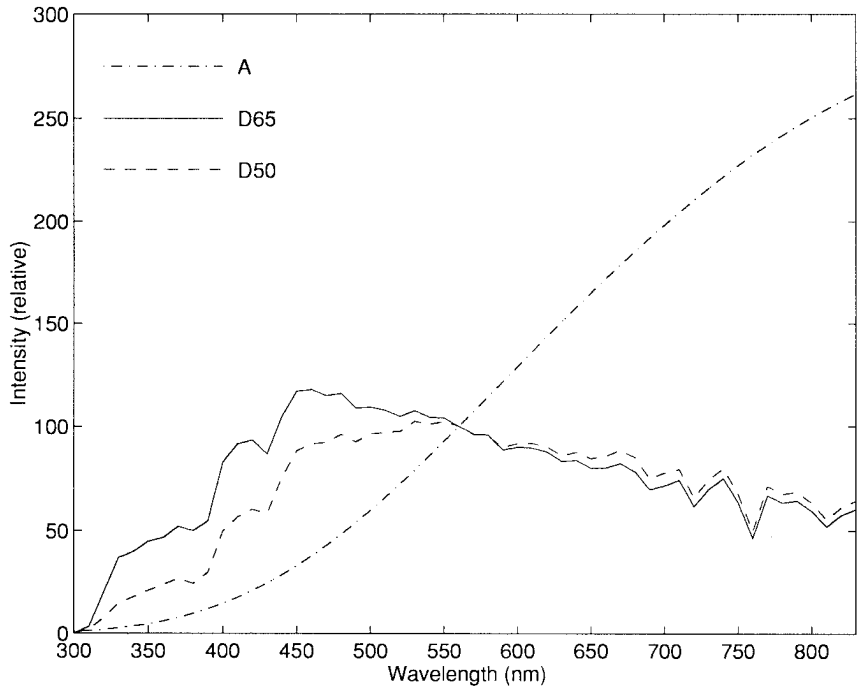

Fig. 5. CIE standard illuminants.

*2) Chromaticity Coordinates and Chromaticity Diagrams:* Since color is specified by tristimuli, different colors may be visualized as vectors in 3-D space. However, such a visualization is difficult to reproduce on two-dimensional (2-D) media and therefore inconvenient. A useful 2-D representation of colors is obtained if tristimuli are normalized to lie in the *unit plane*, i.e., the plane over which the tristimulus values sum up to unity. Such a normalization is convenient as it destroys only information about the "intensity" of the stimulus and preserves complete information about the direction. The coordinates of the normalized tristimulus vector are called *chromaticity coordinates*, and a plot of colors on the unit plane using these coordinates is called a *chromaticity diagram*. Since the three chromaticity coordinates sum up to unity, typical diagrams plot only two chromaticity coordinates along mutually perpendicular axes.

The most commonly used chromaticity diagram is the *CIE xy chromaticity diagram*. The CIE xyz chromaticity coordinates can be obtained from the $X, Y, Z$ tristimulus values in CIE XYZ space as

$$
\begin{aligned}
x &= \frac{X}{X + Y + Z} \\
y &= \frac{Y}{X + Y + Z} \\
z &= \frac{Z}{X + Y + Z}.
\end{aligned}
\tag{9}
$$

Fig. 6 shows a plot of the curve corresponding to visible monochromatic spectra on the CIE xy chromaticity diagram. This shark-fin-shaped curve, along which the wavelength (in nm) is indicated, is called the *spectrum locus*. From the linear relation between irradiance spectra and the tristimulus values, it can readily be seen that the chromaticity coordinates of any additive-combination of two spectra lie on the line segment joining their chromaticity coordinates [12]. From this observation, it follows that the region of chromaticities of all realizable spectral stimuli is the convex hull of the spectrum locus. In Fig. 6, this region of physically realizable chromaticities is the region inside the closed curve formed by the spectrum locus and the broken line joining its two extremes, which is known as the *purple line*.

*3) Transformation of Primaries—NTSC, SMPTE, and CCIR Primaries:* If a different set of primary sources, $\mathbf{Q}$, is used in the color matching experiment, a different set of CMF's, $\mathbf{B}$, are obtained. Since all CMF's are nonsingular linear transformations of the human cone responses, the CMF's are related by a linear transformation. The relation between the two color-matching matrices is given by [37]

$$
\mathbf{B}^T = (\mathbf{A}^T \mathbf{Q})^{-1} \mathbf{A}^T.
\tag{10}
$$

Note that the columns of the $3 \times 3$ matrix $\mathbf{A}^T \mathbf{Q}$ are the tristimulus values of the primaries $\mathbf{Q}$ with respect to the primaries $\mathbf{P}$. Note also that the same transformation, $(\mathbf{A}^T \mathbf{Q})^{-1}$, is useful for the conversion of tristimuli in the primary system $\mathbf{P}$ to tristimuli in the primary system $\mathbf{Q}$.

Color television (TV) was one of the first consumer products exploiting the phenomenon of trichromacy. The three light-emitting color phosphors in the television CRT form the three primaries in this "color-matching experiment." In the United States, the National Television Systems Committee (NTSC) recommendations for a receiver primary system based on three phosphor primaries were adopted by the Federal Communications Commission (FCC) in 1953 for use as a standard in color TV. The FCC standard specified the CIE xy chromaticity coordinates for the phosphors [45] as $(0.670, 0.330)$ (red), $(0.210, 0.710)$ (green), $(0.140, 0.080)$ (blue) [46]. In addition, the tristimulus values $(1, 1, 1)$ were assumed to correspond to a "white color" typically specified as the illuminant D65. The chromaticity coordinates along with the white balance condition define the CIE XYZ tristimuli of the NTSC primaries, which determine the relation of NTSC RGB tristimuli to CIE XYZ tristimuli as per (10).

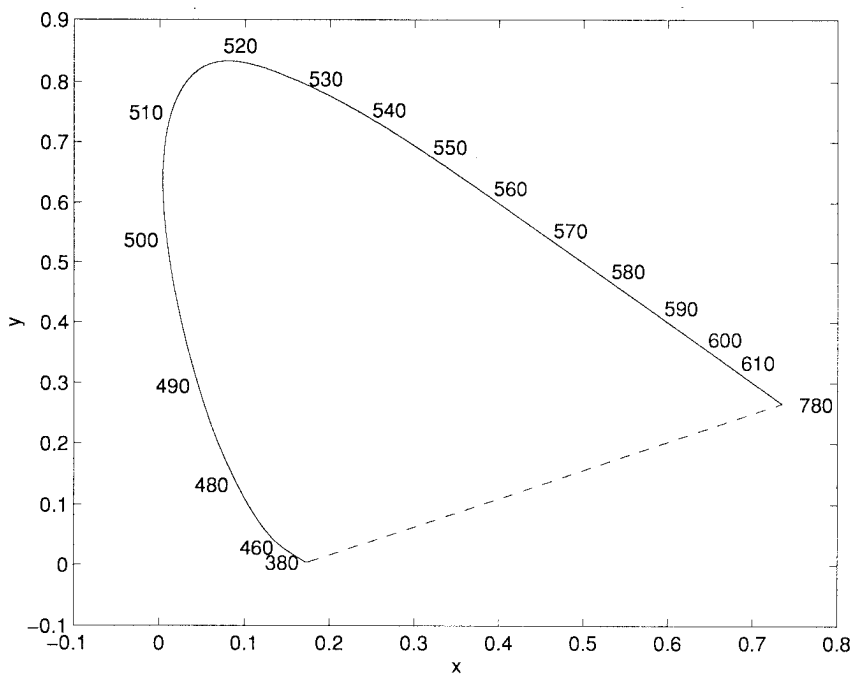

Fig. 6. CIE xy chromaticity diagram.

In the early color TV system, the signal-origination colorimetry was coupled with the colorimetry of displays, with the tacit assumption that the processing at the receiver involves only decoding and no color processing is performed. As display technology changed, manufacturers began using more efficient phosphors and incorporated some changes in the decoding as a compensation for the nonstandard phosphors [47]. Similar changes took place in the monitors used by broadcasters, but they were unaware of the compensating mechanisms in the consumer TV sets. As a result, there was considerable color variability in the broadcast TV system [45]. To overcome this problem, the chromaticities of a set of controlled phosphors were defined for use in broadcast monitors, which now forms the Society of Motion Picture and Television Engineers (SMPTE) "C" phosphor specification [48], [49]. Current commercial TV broadcasts in the United States are based on this specification.

With the development of newer display technologies that are not based on CRT's (see Section III-A4), it is now recognized that signal-origination colorimetry needs to be decoupled from the receiver colorimetry and that color correction at the receiver should compensate for the difference. However, for compatibility reasons and to minimize noise in transformations, it is still desirable to keep the reference primaries for broadcast colorimetry close to the phosphor primaries. Toward this end, the International Radio Consultative Committee (CCIR) [50] has defined a set of phosphor primaries by the chromaticity coordinates $(0.640, 0.330)$ (red), $(0.300, 0.600)$ (green), and $(0.150, 0.060)$ (blue) for use in high definition television (HDTV) systems.

Prior to transmission, tristimuli in SMPTE RGB and CCIR RGB spaces are nonlinearly compressed (by raising them to a power of 0.45) and encoded for reducing transmission bandwidth [50], [51] (the reasons for these operations will be explained in Sections III-A1 and IV-C). Note however, that the encoding and nonlinear operations must be reversed before the signals can be converted to tristimuli spaces associated with other primaries. Transformations for the conversion of color tristimulus values between various systems can be found in [52, pp. 66–67], [53, p. 71], [54], and [55].

### C. Uniform Color Spaces and Color Differences

The standards for colorimetry defined in Section II-B provide a system for specifying color in terms of tristimulus values that can be used to represent colors unambiguously in a 3-D space. It is natural to consider the relation of the distance between colors in this 3-D space to the perceived difference between them. Before such a comparison can be made, it is necessary to have some means for quantifying perceived color differences. For widely different color stimuli, an observer's assessment of the magnitude of color difference is rather variable and subjective [12, p. 486]. At the same time, there is little practical value in quantifying large differences in color, and therefore most research has concentrated on quantifying small color differences. For this purpose, the notion of a *just* *noticeable difference* (JND) in stimuli has been extensively used as a unit by color scientists. An alternate empirically derived system, which has also been used often, is the Munsell color system [56], [57]. In the Munsell system, all possible colors are defined in terms of the perceptual attributes of *lightness, hue,* and *chroma*; and associated Munsell book(s) of color contain reflective samples, which (when viewed under daylight), are spaced apart in perceptually equal steps of these attributes [12]. Lightness, hue, chroma, and other terms of color perception will be used in this paper in accordance with common terminology, but a definition will not be attempted here because of their subjective nature. Definitions are, however, provided in [12, p. 487], [58], [59], and [60].

Several researchers have examined the distribution of JND colors in CIE xy chromaticity and CIE XYZ tristimuli spaces and have found that it varies widely over the color space [61]–[65]. Hence, the CIE XYZ space is perceptually nonuniform in that equal perceptual differences between colors do not correspond to equal distances in the tristimulus space. Since perceptual uniformity is an extremely desirable feature for defining tolerances in color reproduction systems, considerable research has been directed toward the development of *uniform color spaces*. Traditionally, the problem has been decomposed into two sub-problems: i) one of determining a uniform lightness scale, and ii) the other of determining a uniform chromaticity diagram for equilightness color stimuli. The two are then combined with suitable scaling factors for the chromaticity scale and the lightness scale to make their units correspond to the same factor of a JND.

The historical milestones in the search for uniform brightness and lightness scales are described in Wyszecki and Stiles [12, pp. 493–499]. Typical experiments determine these scales either by a process of repeated bisection of the scale extremes or by moving up in increments of a JND. A cube-root power law relation between brightness and luminance provides a satisfactory fit for most experimental data and, therefore, has the most widespread acceptance at present [12, p. 494].

The search for a uniform lightness scale was complemented by efforts toward determination of a *uniform chromaticity scale* for constant lightness. Two of these attempts are noteworthy. The first determined a linear transformation of the tristimulus space that yielded a chromaticity diagram with JND colors being roughly equispaced [66], [67]. This was the precursor of the CIE 1960 u, v diagram [12, p. 503]. The second was primarily motivated by the Munsell system and used a nonlinear transformation of the CIE XYZ tristimuli to obtain a *chromatic-value* diagram in which the distances of Munsell colors of equal lightness would be in proportion to their hue and chroma differences [68]. The form for the nonlinear transformation was based on a color vision model proposed earlier by Adams [69], and the diagram is therefore referred to as Adams' chromatic-value diagram.

Based on the aforementioned research, the CIE has recommended two uniform color spaces for practical applications: the CIE 1976 $L^*u^*v^*$ (CIELUV) space and the CIE 1976 $L^*a^*b^*$ (CIELAB) space [40]. These spaces are defined in terms of transformations from CIE XYZ tristimuli into these spaces. Both spaces employ a common lightness scale, $L^*$, that depends only on the luminance value $Y$. The lightness scale is combined with different uniform chromaticity diagrams to obtain a 3-D uniform color space. For the CIELUV space, a later version of CIE 1960 u, v diagram is used whereas CIELAB uses a modification of Adams' chromatic-value diagram [12, p. 503]. In either case, the transformations include a normalization involving the tristimuli of a white stimulus, which provides a crude approximation to the eye's adaptation (see Section II-D1). Euclidean distances in either space provide a color-difference formula for evaluating color differences in perceptually relevant units.

*1) The CIE 1976 $L^*u^*v^*$ Space:* The $L^*, u^*, v^*$ values corresponding to a stimulus with CIE XYZ tristimulus values $X, Y, Z$ are given by [40]

$$L^* = 116 f\left(\frac{Y}{Y_n}\right) - 16 \tag{11}$$

$$u^* = 13 L^*(u' - u'_n) \tag{12}$$

$$v^* = 13 L^*(v' - v'_n) \tag{13}$$

where

$$f(x) = \begin{cases} x^{\frac{1}{3}} & x > 0.008856 \\ 7.787x + \frac{16}{116} & x \leq 0.008856 \end{cases} \tag{14}$$

$$u' = \frac{4X}{X + 15Y + 3Z} \tag{15}$$

$$v' = \frac{9Y}{X + 15Y + 3Z} \tag{16}$$

$$u'_n = \frac{4X_n}{X_n + 15Y_n + 3Z_n} \tag{17}$$

$$v'_n = \frac{9Y_n}{X_n + 15Y_n + 3Z_n} \tag{18}$$

and $X_n, Y_n, Z_n$ are the tristimuli of the white stimulus.

The Euclidean distance between two color stimuli in CIELUV space is denoted by $\triangle E^*_{uv}$ (delta E-uv), and is a measure of the total color difference between them. On an average, a $\triangle E^*_{uv}$ value of around 2.9 corresponds to a JND [70]. As mentioned earlier, the value of $L^*$ serves as a correlate of lightness. In the $u^*v^*$ plane, the radial distance $(\sqrt{(u^*)^2 + (v^*)^2})$ and angular position $(\arctan \frac{u^*}{v^*})$ serve as correlates of chroma and hue, respectively.

*2) The CIE 1976 $L^*a^*b^*$ Space:* The $L^*$ coordinate of the CIELAB space is identical to the $L^*$ coordinate for the CIELUV space, and the transformations for the $a^*$ and $b^*$ coordinates are given by

$$a^* = 500\left(f\left(\frac{X}{X_n}\right) - f\left(\frac{Y}{Y_n}\right)\right) \tag{19}$$

$$b^* = 200\left(f\left(\frac{Y}{Y_n}\right) - f\left(\frac{Z}{Z_n}\right)\right) \tag{20}$$

where $f(\cdot), X_n, Y_n$, and $Z_n$ are as defined earlier.

The Euclidean distance between two color stimuli in CIELAB space is denoted by $\triangle E^*_{ab}$ (delta E-ab), and a $\triangle E^*_{ab}$ value of around 2.3 corresponds to a JND [70]. Once again, in the $a^*b^*$ plane, the radial distance $(\sqrt{(a^*)^2 + (b^*)^2})$ and angular position $(\arctan \frac{a^*}{b^*})$ serve as correlates of chroma and hue, respectively.

*2) Other Color Difference Formulae:* As may be expected, the CIELUV and CIELAB color spaces are only approximately uniform and are often inadequate for specific applications. The uniformity of CIELAB and CIELUV is about the same, but the largest departures from uniformity occur in different regions of the color space [71]–[73]. Several other uniform color spaces and color difference formulae have been proposed since the acceptance of the CIE standards. Since CIELAB has gained wide acceptance as a standard, most of the difference formulae attempt to use alternate (non-Euclidean) "distance measures"[1] in the CIELAB space. Prominent among these are the CMC (l:c) distance function based on the CIELAB space [74] and the BFD (l:c) function [75], [76]. A comparison of these and other uniform color spaces using perceptibility and acceptability criteria appears in [70]. In image processing applications involving color, the CIELAB and CIELUV spaces have been used extensively, whereas in industrial color control applications the CMC formulae have found wider acceptance. Recently [77], the CIE issued a new recommendation for the computation of color differences in CIELAB space that incorporates several of the robust and attractive features of the CMC (l:c) distance function.

### D. Psychophysical Phenomena and Color Appearance Models

The human visual system as a whole displays considerable adaptation. It is estimated that the total intensity range over which colors can be sensed is around $10^8 : 1$. While the cones themselves respond only over a $1000 : 1$ intensity range, the vast total operating range is achieved by adjustment of their sensitivity to light as a function of the incident photon flux [78]. This adjustment is believed to be largely achieved through a feedback from the neuronal layers that provide temporal lowpass filtering and adjust the cones output as a function of average illumination. A small fraction of the adaptation corresponding to a factor of around $8 : 1$ is the result of a $4 : 1$ change in the diameter of the pupil that acts as the aperture of the eye [60, p. 23].

Another fascinating aspect of human vision is the invariance of object colors under lights with widely varying intensity levels and spectral distributions. Thus objects are often recognized as having approximately the same color in phases of daylight having considerable difference in their spectral power distribution and also under artificial illumination. This phenomenon is called *color constancy*. The term *chromatic adaptation* is used to describe the changes in the visual system that relate to this and other psychophysical phenomena.

While colorimetry provides a representation of colors in terms of three independent variables, it was realized early on that humans perceive color as having four distinct *hues* corresponding to the perceptually unique sensations of red, green, yellow, and blue. Thus, while yellow can be produced by the additive combination of red and green, it is clearly perceived as being qualitatively different from each of the two components. Hering [79] had considerable success in explaining color perception in terms of an *opponent-colors theory*, which assumed the existence of neural signals of opposite kinds with the red–green hues forming one opponent pair and the yellow–blue hues constituting the other. Such a theory also satisfactorily explains both the existence of some intermediate hues (such as red–yellow, yellow–green, green–blue, and blue–red) and the absence of other intermediate hues (such as reddish-greens and yellowish-blues).

Initially, the trichromatic theory and the opponent-colors theory were considered competitors for explaining color vision. However, neither one by itself was capable of giving satisfactory explanations of several important color vision phenomena. In more recent years, these competing theories have been combined in the form of *zone theories of color vision*, which assume that there are two separate but sequential zones in which these theories apply. Thus, in these theories it is postulated that the retinal color sensing mechanism is trichromatic, but an opponent-color encoding is employed in the neural pathways carrying the retinal responses to the brain. These theories of color vision have formed the basis of a number of color appearance models that attempt to explain psychophysical phenomena. Typically in the interests of simplicity, these models follow the theories only approximately and involve empirically determined parameters. The simplicity, however, allows their practical use in color reproduction applications involving different media where a perceptual match is more desirable and relevant than a colorimetric match.

A somewhat different but widely publicized color vision theory was the *retinex* (from *retin*a and cort*ex*) theory of Edwin Land [80], [81]. Through a series of experiments, Land demonstrated that integrated broadband reflectances in red, green, and blue channels show a much stronger correlation with perceived color than the actual spectral composition of radiant light incident at the eye. He further postulated that the human visual system is able to infer the broadband reflectances from a scene through a successive comparison of spatially neighboring areas. As a model of human color perception, the retinex theory has received only limited attention in recent literature, and has been largely superseded by other theories that explain a wider range of psychophysical effects. However, a computational version of the theory has recently been used, with moderate success, in the enhancement of color images [82], [83].

One may note here that some of the uniform color spaces include some aspects of color constancy and color appearance in their definitions. In particular, both the CIELAB and CIELUV spaces employ an opponent-color encoding and use white-point normalizations that partly explain color constancy. However, the notion of a color appearance model is distinct from that of a uniform color space. Typical uniform color spaces are useful only for comparing stimuli under similar conditions of adaptation and can yield incorrect results if used for comparing stimuli under different adaptation conditions.

*1) Chromatic Adaptation and Color Constancy:* Several mechanisms of chromatic adaptation have been proposed to explain the phenomenon of color constancy. Perhaps the most widely used of these in imaging applications is one proposed by Von Kries [84]. He hypothesized that the chromatic adaptation is achieved through individual adaptive gain control on each of the three cone responses. Thus, instead of (2), a more complete model represents the cone responses as

$$\mathbf{c}' = \mathbf{D}\mathbf{S}^T\mathbf{f} \qquad (21)$$

where $\mathbf{D}$ is a diagonal matrix corresponding to the gains of the three channels, and the other terms are as before. The gains of the three channels depend on the state of adaptation of the eye, which is determined by preexposed stimuli and the surround, but independent of the test stimulus $\mathbf{f}$. This is known as the *Von Kries coefficient rule*.

---

[1] Note several of these distance measures are asymmetric and as such do not satisfy the mathematical requirements for a metric [22, p. 91].

The term *asymmetric-matching* is used to describe matching of color stimuli under different adaptation conditions. Using the Von Kries coefficient rule, two radiant spectra, $\mathbf{f}_1$ and $\mathbf{f}_2$, viewed under adaptation conditions specified by the diagonal matrices, $\mathbf{D}_1$ and $\mathbf{D}_2$, respectively, will match if

$$\mathbf{D}_1\mathbf{S}^T\mathbf{f}_1 = \mathbf{D}_2\mathbf{S}^T\mathbf{f}_2. \qquad (22)$$

Thus, under the Von Kries coefficient rule, chromatic adaptation can be modeled as a diagonal transformation for tristimuli specified in terms of the eye's cone responses. Usually, tristimulus values are specified not relative to the cone responses themselves, but to CMF's that are linear transformations of the cone responses. In this case, it can readily be seen [12, p. 432] that the tristimuli of color stimuli that are in an asymmetric color match are related by a similarity transformation [85] of the diagonal matrix $\mathbf{D}_1^{-1}\mathbf{D}_2$.

A Von Kries transformation is commonly used in color rendering applications because of its simplicity and is a part of several standards for device-independent color imaging [86], [87]. Typically, the diagonal matrix $\mathbf{D}_1^{-1}\mathbf{D}_2$ is determined by assuming that the cone responses on either side of (22) are identical for white stimuli (usually a perfect reflector illuminated by the illuminant under consideration). The white-point normalization in CIELAB space was primarily motivated by such a model. Since the CIE XYZ CMF's are not *per se* the cone responses of the eye, the diagonal transformation representing the normalization is not a Von Kries transformation and was chosen more for convenience than accuracy [88].

In actual practice, the Von Kries transformation can explain results obtained from psychophysical experiments only approximately [12, pp. 433–451]. At the same time, the constancy of metameric matches under different adaptation conditions provides strong evidence for the fact that the cone response curves vary only in scale while preserving the same shape [89, p. 15]. Therefore, it seems most likely that part of the adaptation lies in the nonlinear processing of the cone responses in the neural pathways leading to the brain.

A number of alternatives to the Von Kries adaptation rule have been proposed to obtain better agreement with experimental observations. Most of these are nonlinear and use additional information that is often unavailable in imaging applications. A discussion of these is beyond the scope of this paper, and the reader is referred to [60, pp. 81, 217], [90]–[92], and [88] for examples of such models.

The phenomenon of color constancy suggests that the human visual system transforms recorded stimuli into representations of the scene reflectance that are (largely) independent of the viewing illuminant. Several researchers have investigated algorithms for estimating illuminant-independent descriptors of reflectance spectra from recorded tristimuli, which have come to be known as *computational color constancy algorithms* [93]–[97]. Several of these algorithms rely on low-dimensional linear models of object and illuminant spectra, which will be discussed briefly in Section III-B5. A discussion of how these algorithms relate to the Von Kries transformation rule and to human color vision can also be found in [98], [95], and [97].

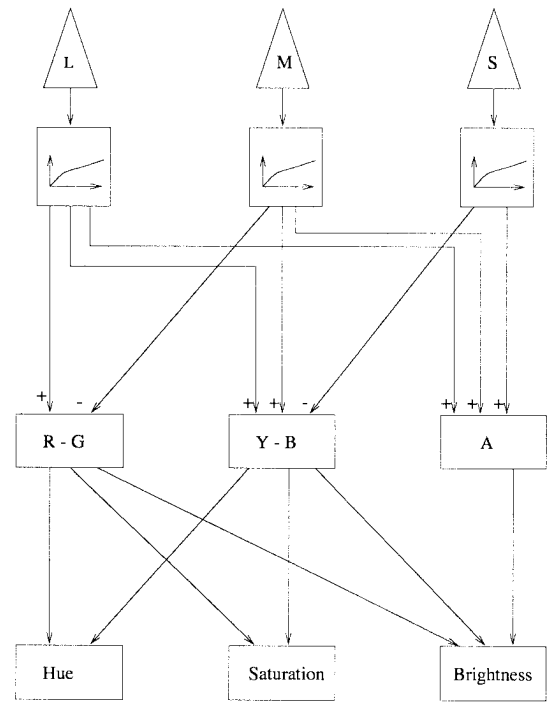

Fig. 7. Typical "wiring diagram" for human color vision models (adapted from [99]).

*2) Opponent Processes Theory and Color Appearance Models:* The modeling of chromatic adaptation is just one part of the overall goal of color appearance modeling. While color appearance models are empirically determined, they are usually based on physiological models of color vision. Most modern color vision models are based on "wiring diagrams" of the type shown in Fig. 7. The front end of the model consists of L, M, and S (long, medium, and short wavelength sensitive) cones. The cone responses undergo nonlinear transformations and are combined into two opponent color chromatic channels (R-G and Y-B), and one achromatic channel (A). A positive signal in the R-G channel is an indication of redness whereas a negative signal indicates greenness. Similarly, yellowness and blueness are opposed in the Y-B channel. The outputs of these channels combine to determine the perceptual attributes of hue, saturation, and brightness.

It is obvious that the above color-vision model is an over simplification. Actual color appearance models are considerably more intricate and involve a much larger number of parameters, with mechanisms to account for spatial effects of surround and the adaptation of the cone responses, which was briefly discussed in the last section. Due to the immense practical importance of color appearance modeling to color reproduction systems, there has been considerable research in this area that cannot be readily summarized here. The interested reader is referred to [60, pp. 213–258] and [100]–[112] for examples of some of the prominent color appearance models in current literature. A recent overview of the current understanding of human color vision can also be found in [113].

## III. COLOR REPRODUCTION AND RECORDING SYSTEMS

The basics of color discussed in the last section addressed the issue of specification of a single color stimulus. In practical systems, one is usually concerned with the processing of color images with a large number of colors. In the physical world, these images exist as spatially varying spectral radiance or reflectance distributions. Color information needs to be recorded from these distributions before any processing can be attempted. Conversely, the physical realization of color images from recorded information requires synthesis of spatially varying spectral radiance or reflectance distributions. In this section, some of the common color output and input systems are surveyed. Output systems are discussed first because color recording systems may also be used to record color reproductions and may exploit the characteristics of the reproduction device.

### A. Color Output Systems

Nature provides a variety of mechanisms by which color may be produced. As many as fifteen distinct physical mechanisms have been identified that are responsible for color in nature [114]. While only a fraction of these mechanisms is suitable for technological exploitation, there is still considerable diversity in available technologies and devices for displaying and printing color images.

Color output devices can broadly be classified into three types: *additive*, *subtractive*, and *hybrid*. Additive color systems produce color through the combination of differently colored lights, known as primaries. The qualifier additive is used to signify the fact that the final spectrum is the sum (or average) of the spectra of the individual lights, as was assumed in the discussion of color matching in Section II-A1. Examples of additive color systems include color CRT displays and projection video systems. Color in subtractive systems is produced through a process of removing (subtracting) unwanted spectral components from "white" light. Typically, such systems produce color on transparent or reflective media, which are illuminated by white light for viewing. Dye sublimation printers, color photographic prints, and color slides are representatives of the subtractive process. Hybrid systems use a combination of additive and subtractive processes to produce color. The main use of a hybrid system is in color halftone printing, which is commonly used for lithographic printing and in most desktop color printers.

Any practical output system is capable of producing only a limited range of colors. The range of producible colors on a device is referred to as its *gamut*. The gamut of a device is a 3-D object and can be visualized using a 3-D representation of the color space, such as the colorimetry standards or uniform color spaces discussed earlier [115], [116]. Often, 2-D representations are more convenient for display, and chromaticity diagrams are used for this purpose. From the linearity of color matching, it can be readily seen that the gamut of additive systems in CIE XYZ space (or any of the other linear tristimulus spaces) is the convex polyhedron formed by linear combinations of the color tristimuli of the primaries over the realizable amplitude range. On the CIE xy chromaticity diagram, the gamut appears as a convex polygon with the primaries representing the vertices. For the usual case of three red, green, and blue primaries, the gamut appears as a triangle on the CIE xy chromaticity diagram. Since most subtractive and hybrid systems are nonlinear, their gamuts have irregular shape and are not characterized by such elegant geometric constructs. One may note here that in order to obtain the largest possible chromaticity gamut, most three-primary additive systems use red, green, and blue colored primaries. For the same reason, cyan, magenta, and yellow primaries are used in subtractive and hybrid systems.

In order to discuss colorimetric reproduction on color output devices, it is useful to introduce some terminology. The term *control values* is used to denote signals that drive a device. The operation of the device can be represented as a multidimensional mapping from control values to colors specified in a device-independent color space. This mapping is referred to as the (device) *characterization*. Since specified colors in a device-independent color space need to be mapped to device control values to obtain colorimetric output, it is necessary to determine the inverse of the multidimensional device-characterization function. In this paper, the term *calibration* will be used for the entire procedure of characterizing a device and determining the inverse transformation. If the device's operation can be accurately represented by a parametric model, the characterization is readily done by determining the model parameters from a few measurements. If no useful model exists, a purely empirical approach is necessary, in which the characterization function is directly measured over a grid of device control values. The inversion may be performed in a closed form if the characterization uses a device model that allows this. If an empirical approach is employed in characterization or if the model used is noninvertible (often the case with nonlinear models), one has to resort to numerical methods in the inversion step.

*1) Cathode Ray Tubes:* The most widely used display device for television and computer monitors is the color CRT. The CRT produces visible light by bombardment of a thin layer of phosphor material by an energetic beam of electrons. The electron beam causes the phosphor to fluoresce and emit light whose spectral characteristics are governed by the chemical nature of the phosphor. The most commonly used color CRT tubes are the shadow-mask type, in which a mosaic of red, green, and blue light emitting phosphors on a screen is illuminated by three independent electron beams. The intensity of light emitted by the phosphors is governed by the velocity and number of electrons. The beam is scanned across the screen by electrostatic or electromagnetic deflection mechanisms. The number of electrons is modulated in synchronism with the scan to obtain spatial variations in the intensity of the light emitted by the three phosphors. At normal viewing distances, the light from the mosaic is spatially averaged by the eye, and the CRT thus forms an additive color system.

There are several design choices in the manufacture of shadow mask CRT's. Other competing designs offer improved resolution by utilizing a layered phosphor instead of a mosaic. The reader is referred to [117] and [118] for a description of the different technologies and involved tradeoffs. A detailed description of physical principles involved in the operation of these devices is provided in [119, pp. 79–200].

Color in CRT displays is controlled through the application of different voltages to the red, green, and blue guns. For a complete colorimetric characterization of these devices, the CIE XYZ tristimulus values (or other tristimuli) need to be specified as a spatially varying function of the voltages applied to the three guns. A brute force approach to this problem, using a multidimensional look-up table, is infeasible because of the extremely large number of measurements required. Hence, simplifying assumptions need to be made in order to make the problem tractable.

Assumptions of spatial uniformity, gun independence, and phosphor constancy are commonly made in order to simplify CRT colorimetry [120]. Spatial uniformity implies that the color characterization of the CRT does not vary with position. Gun independence refers to the assumption that the three phosphors and their driving mechanisms do not interact. This implies that the incident intensity at the eye when the guns are operated simultaneously is the sum of the intensities when the guns are operated individually. Phosphor constancy refers to the assumption that the relative spectral power distribution of light emitted by the phosphors does not change with driving voltage (i.e., at all driving voltages the spectra emitted by a phosphor are scalar multiples of a single spectrum).

With the above three assumptions, the problem of characterizing the CRT reduces to a problem of relating the amplitudes of the individual red, green, and blue channels to their corresponding gun voltages. The problem can be further simplified through the use of a parametric model for the operation of the individual guns. Typically, these models are based on the exponential relation between the beam current and grid voltage in vacuum tubes [121], [122]. For each gun, the spectrum of emitted light in response to a control voltage, $v$, is modeled by an expression of the form $(v/v_m)^\gamma h(\lambda)$, where $v_m$ is the maximum value of the voltage, $h(\lambda)$ is the emitted phosphor spectrum at the maximum voltage, and $\gamma$ is the exponential parameter. The exponent, $\gamma$, is commonly referred to as the *monitor-gamma* and is normally around 2.2 for most color monitors. Since the above parametric model is only approximate, several modifications of it have been used by researchers [123]–[126]. Using the parametric models, CRT monitors can be readily characterized using a small number of measurements.

In order to produce colorimetric color on a CRT display, the "inverse" of the characterization, i.e., the transformation from CIE XYZ tristimuli to the driving voltages for the guns, is required. Since the characterization is on a per-channel basis, the transformation from CIE XYZ tristimulus values can also be determined as a linear transformation, corresponding to a transformation from the CIE primaries to the phosphor primaries, followed by a one-dimensional (1-D) transformation that is determined by the parametric model used to represent the operation of the individual electron guns [125]. Typically, this operation involves exponentiation to the power of $1/\gamma$ and is known as *gamma-correction*. As mentioned in Section II-B3, TV signals are normally gamma corrected before transmission. One may note here that quantization of gamma corrected signals results in wider quantization intervals at higher amplitudes where the sensitivity of the eye is also lower. Therefore, just like speech companding, gamma correction of color tristimuli prior to quantization in a digital system (or transmission in a limited bandwidth system) reduces the perceptibility of errors and contours in comparison to a scheme in which no gamma correction is used [73], [127]–[130, p. 393].

For colors that the phosphors are capable of producing, fairly good color reproduction can be obtained on a CRT using the models mentioned above. Berns *et al.* [125] report an accuracy around 0.4 $\triangle E_{ab}^*$ using only eight measurements for determining model parameters. However, the gamut of CRT tubes is limited by the phosphors used, which causes significant color errors for colors that lie beyond the gamut. This is one of the primary sources of color errors seen in broadcast TV.

The assumptions of gun independence and phosphor constancy have been tested by several researchers and found to hold reasonably well [123], [131], [120], [126]. However, in most CRT monitors for the same driving voltage, the light intensity is brightest at the center and falls off toward the edges. The change in luminance over the screen can be as high as 25% [132, p. 104]. Therefore, the assumption of spatial uniformity does not strictly hold. Since the eye's sensitivity is not uniform over the entire field of view and because the eye adapts well to the smooth variation in intensity across the screen, the spatial nonuniformity of CRT's is not too noticeable. An algorithm for correcting for spatial inhomogeneity is discussed in [133].

*2) Contone Printers:* Continuous tone ("contone") printers use subtractive color reproduction to produce color images on (special) paper or transparent media. Subtractive color reproduction is achieved by using cyan, magenta, and yellow colorants in varying concentrations to absorb different amounts of light in the red, green, and blue spectral regions, respectively. Each colorant absorbs its complimentary color and transmits the rest of the spectrum. The extent of absorption is determined by the concentration of the colorant, and the use of different concentrations produces different colors. For an excellent description of the subtractive process and the reasons behind the choice of cyan, magenta, and yellow colorants, the reader is referred to [134, Chap. 3].

The subtractive principle is schematically shown in Fig. 8 for a transmissive system. If the incident light spectrum is $l(\lambda)$, the spectrum of the light transmitted through the three layers is given by, $g(\lambda) = l(\lambda)t_1(\lambda)t_2(\lambda)t_3(\lambda)$, where $t_i(\lambda)$ is the spectral transmittance of the $i$th layer. If the colorants are transparent (i.e., do not scatter incident light) and their absorption coefficients are assumed to be proportional to their concentration (Bouguer–Beer law), it can be shown [135, Chap. 7] that the *optical density* of the $i$th colorant layer, which is defined as the logarithm (base 10) of its transmittance, is given by

$$d_i(c_i; \lambda) = \log_{10} t_i(c_i; \lambda) = c_i d_i(\lambda) \qquad (23)$$

where $t_i(c_i; \lambda)$ is the transmittance of the $i$th colorant layer, $c_i$ is the concentration of the $i$th colorant which varies between 0 and 1, and $d_i(\lambda) = d_i(1; \lambda)$ is the density at maximum concentration.

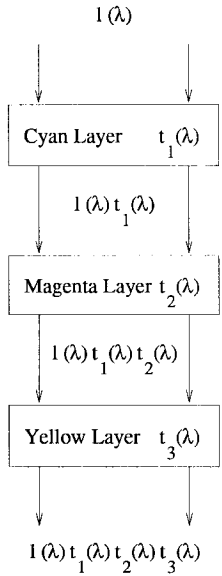

Fig. 8. Subtractive color reproduction.

Using samples of the spectra involved, the spectrum of transmitted light can be represented as [38]

$$\mathbf{g} = \mathbf{L}[10^{-\mathbf{Dc}}] \qquad (24)$$

where $\mathbf{L}$ is a diagonal matrix representing an illuminant spectrum, $\mathbf{c} = [c_1 c_2 c_3]^T$ is the vector of colorant concentrations, $\mathbf{D} = [\mathbf{d_1 d_2 d_3}]$, the remaining boldface symbols represent sample vectors of the corresponding spectral functions, and the exponentiation is computed componentwise.

For prints produced on paper, the transmitted light is reflected by the paper surface and travels once again through the colorant layers. This process is readily incorporated in the model of (24) as an additional diagonal matrix that represents the reflectance spectrum of the substrate and a doubling of the densities $\{d_i(\lambda)\}_{i=1}^3$. For simplicity, the substrate reflectance can be conceptually included in the illuminant matrix $\mathbf{L}$ and the same equations can be used for reflective media.

Even the simplified model of (24) cannot be used for a closed-form calibration of a subtractive system. Analytical models therefore often assume that the three dyes have nonoverlapping rectangular shaped absorptance functions. This is known as the *block dye* assumption. Using the block dye assumption, colorant concentrations required to produce a given CIE tristimulus can be determined in closed form [38].

Most contone printers available currently use thermal dye diffusion technology. The coloring dyes in such a system are transferred from a dispensing film into the reproduction medium by means of heat-induced diffusion. Often, a fourth black dye is used in addition to the cyan, magenta, and yellow dyes to achieve greater contrast and better reproduction of achromatic (gray) colors. A recent review of the physical processes involved in a thermal dye diffusion printer can be found in [136] and [137]. Note that conventional photography uses subtractive color reproduction, thus continuous tone printing is possible using photochemical methods that mimic photography. For an overview of this method and for more details on the dye diffusion printing process, the reader is referred to [138] and [139].

In practice, the models for subtractive printers described above do not provide even reasonable approximations for actual printers. The colorants have smooth absorptance curves that do not agree with the block dye assumption. In addition, typical colorants are not completely transparent, and therefore the Kubelka–Munk theory [140], [135], [141], which accounts for scattering of light by the colorants, is more appropriate instead of the Bouguer–Beer law. These modifications have successfully been used in [142] to model a thermal dye diffusion printer. Since accurate determination of the model parameters is fairly complicated and there are interactions between the media and the colorant layers that are not accounted for even in the Kubelka–Munk theory, often purely empirical techniques are used to characterize color printers. At the same time, note that the model in (24), though somewhat restrictive, has proven very useful in analytical simulations of printers and in making design choices [143].

Typical empirical approaches for color printer calibration begin by measuring the color of test prints corresponding to a uniform grid of control values. This provides a sampling of the mapping from device control values to a device-independent color space. A variety of interpolation based techniques are then utilized to determine the required inverse transformation—typically, in the form of a look-up table over a uniform grid in a color space [144]–[146]. Interesting alternate approaches have also utilized neural networks [147] and an iterated scheme that concentrates measurements in regions of greatest impact [148].

*3) Halftone Printers:* Contone printers require reliable and accurate spatial control of colorant concentrations, which is difficult to achieve. As a result, contone printers are rather expensive. Most desktop printers are therefore based on the simpler technique of halftoning, which has long been the color reproduction method of choice in commercial lithographic printing. Like CRT displays, halftoning exploits the spatial lowpass characteristics of the human visual system. Color halftone images are produced by placing a large number of small differently colored dots on paper. Due to the lowpass nature of the eye's spatial response, the effective spectrum seen by the eye is the average of the spectra over a small angular subtense. Different colors are produced by varying the relative areas of the differently colored dots. In contrast with contone printing, the concentration of a colorant within a dot is not varied and therefore halftone printers are considerably easier and cheaper to manufacture. Special processing of images is necessary to determine the dot patterns for the different colors prior to printing on a halftone printer. This processing is the subject of Section IVB, and only the halftone printing mechanism will be discussed in this section.

In order to obtain a reasonable gamut, most three-ink halftone systems use cyan, magenta, and yellow colorants for printing the dots [134, Chap. 3]. Just as in contone printers, a fourth black colorant is often introduced to conserve the more expensive colorants, reduce ink usage, and produce denser blacks [134, p. 282]. The colorants combine subtractively over the regions in which they overlap producing up to $2^K$ different colors with $K$ colorants. These distinct colors are called the *Neugebauer primaries* after H. E. J. Neugebauer who first suggested that halftone reproduction may be viewed as an additive process involving these primaries [149].

In Neugebauer's model for halftone printers, the spectral macroreflectance of a halftoned region can be expressed as the weighted average of the reflectances of the individual Neugebauer primaries, with the weighting factor of each primary given by its relative area. The term *macroreflectance* is used to indicate that it is actually an average of an inhomogeneous region of differing (micro)reflectances. Since the model is linear in the reflectances of the Neugebauer primaries, the same weighted average formulation applies to colors specified in a tristimulus space instead of the spectra. Since the original Neugebauer model used a tristimulus space, recent spectral versions of the statement are referred to as the *spectral Neugebauer model* [150]. For a three colorant printer, Demichel [151] suggested a statistical scheme (assuming random coverage) for determining the areas of the Neugebauer primaries from the physical printing areas of the three colorants. As a further simplification, the reflectances of the Neugebauer primaries composed of overprints of more than one colorant may be expressed in terms of the transmittances of the different colorant layers as was done in the subtractive model of (24). However, since this assumption of additivity of densities reduces accuracy, it is usually not invoked.

Prior to the work of Neugebauer, halftone color reproduction was often confused with subtractive reproduction, and the Neugebauer model therefore offered very significant improvements [152]. However, the actual halftone process is considerably more complicated. Due to the penetration and scattering of light in paper, known as the *Yule–Nielsen effect*[2] [153], [154], the simple Neugebauer model does not perform well in practice. As a result, several empirical modifications have been suggested for the model. The papers in [155] are an excellent repository of information on the Neugebauer model and its modifications. More recently, considerable success has been demonstrated in using a spectral Neugebauer model with empirical corrections for the Yule–Nielsen effect [150], [156]. Complete and accurate physical models for the color halftone printing process and the Yule–Nielsen effect continue to be elusive, though recent research [157] has yielded encouraging results.

One obstacle in the direct use of Neugebauer models in characterizing a halftone printer is that the relation between the control values and the printing area of the different colorants is usually not known *a priori*. Hence, an empirical component is normally required even for characterization schemes using a model. This empirical component is in the form of 1-D pretransformations of device control values, which often serve the additional purpose of increasing characterization-accuracy along the *achromatic* or *neutral gray* axis, where the eye has significantly greater sensitivity [158]. Purely empirical schemes similar to those used for characterizing contone printers can also be used for halftone printers. The models mentioned above are nonlinear and nonseparable in the device control values and cannot be inverted analytically. Hence, for both model-based and empirical schemes, the inversion of the characterization needs to be performed numerically. In either case, the final mapping from required color tristimuli to device control values is realized as a multidimensional look-up table. The models, however, have an advantage over a purely empirical approach in that they offer a significant reduction in the number of measurements required. An interesting generalization of the Neugebauer model is discussed in [159] and [160], where the model is interpreted as interpolating between a number of end-points specified by the primaries. Accuracy is improved by using local interpolation over smaller cells, which in turn implies more measurements. The generalization, known as the *cellular Neugebauer model*, thus offers a graceful tradeoff between accuracy and the number of measurements required. Due to the large number of effects ignored by most models, they can offer only limited accuracy. Therefore, in graphic arts and printing industries, where there is greater emphasis on quality, measurement-intensive empirical schemes are often preferred [161]. A comparison of some model-based and measurement-based empirical schemes for electronic imaging applications can be found in [162].

Halftone printers have been manufactured using very different technologies for printing dots on paper [139, pp. 4–8]. The most promising current technologies utilize inkjet, thermal transfer, and electrophotography to produce the halftone dots. Even a brief mention of the principles and technology of these devices is beyond the scope of this paper, and the interested reader is referred to [138], [163], [139] and [164] for details.

*4) Recent Advances in Color Displays and Printing:* The increasing use of portable computers has fostered considerable research in displays that overcome the CRT's problems of bulk, weight, and high power consumption. Active and passive color liquid-crystal displays (LCD's) are already in use in notebook personal computers, and their use is also proposed in wall-mounted displays for HDTV [165]. A number of other technologies, including color light-emitting diodes (LED's), electro-luminescent displays, and plasma displays, are also being actively investigated. A description of their historical development, physical principles, and relative merits and demerits can be found in [118], [119], [166], and [167]. Most of them are additive color systems similar to a CRT and use a mosaic of red, green, and blue "dots" to produce color, though there are also some LCD devices based on the subtractive principle [165], [167] or on spectrally selective reflection [168].

---

[2] Note that in the printing of the original paper [153], Nielsen's name was misspelled as "Neilsen". Both spellings have therefore been used in existing literature.

A majority of the color display devices mentioned so far rely on the spatial lowpass characteristics of the human eye to produce different colors using a mosaic of differently colored regions. An alternative system for producing color, known as *field sequential color* (FSC), is based on the temporal lowpass nature of the eye's response. In an FSC system, red, green, and blue image frames are projected in rapid succession onto the viewing screen, and the temporal averaging in the observer's eye produces the illusion of a single colored image. An FSC system was originally selected by the FCC for color TV transmission, but before it could be commercialized it was replaced by the monochrome-compatible NTSC system in use today. The primary drawback in such a system was the high frame rate and bandwidth requirements [169, pp. 218–219]. Recently, there has been a resurgence of interest in FSC systems. An interesting example of a recent FSC system is Texas Instrument's digital micromirror device (DMD) [170] that utilizes an array of deformable micromirrors. In the deformed state, each micromirror deflects light from an illuminating lamp onto a single picture element (pixel) on the viewing screen. The duty cycles of the deformation of different mirrors are modulated to produce image intensity variations on the screen. Color is produced by placing a color filter-wheel between the lamp and the micromirror device and synchronizing the red, green, and blue frames with the color wheel. Alternate configurations using three separate DMD devices or two devices in a five primary projection system have also been reported [171]. From a color imaging standpoint, DMD displays are rather interesting, as they are almost linear and allow considerable flexibility in the choice of the primaries through the use of different color filters in the filter-wheel.

There have also been significant new advances in color printing. Color halftone printers have continually improved in resolution, speed, and cost. Some devices now incorporate limited contone capability through a coarse variation in colorant concentrations. The gamut of printers has also been enlarged by using improved colorants, or more recently, by using more than three/four inks, which is referred to as *high-fidelity* ("hi-fi") printing [172]–[174].

Since most of the devices mentioned above are still in their infancy, little information if any is available on the color characterization and performance of these devices. As they find increased acceptance, there will no doubt be a greater demand for more accurate color characterization and for reasonable models of these devices. This will, therefore, be an active area of color imaging research in the future.

### B. Color Input Systems

In order to process images digitally, the continuous-space, analog, real-world images need to be sampled and quantized. Requirements for the spatial sampling process and the effects of quantization have been analyzed in considerable detail in signal processing and communications literature and will not be reiterated here. This section will, instead, look at the requirements of devices that sample spectral information at each spatial location and attempt to preserve color information.

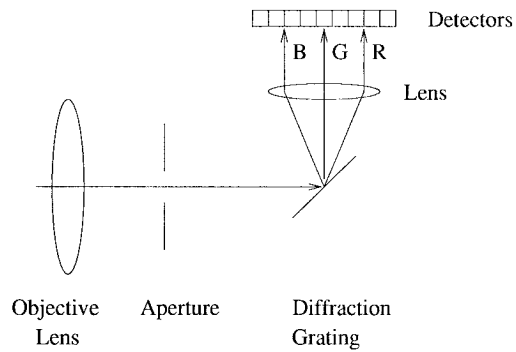

Fig. 9. Schematic cross section of a typical spectroradiometer.

*1) Spectroradiometers and Spectrophotometers:* Sampling of the spectral distribution provides the most direct and complete technique for recording color information. A *spectroradiometer* is the fundamental device used to record the spectrum. Fig. 9 shows the schematic cross section of a spectroradiometer. The light is collimated by the lens onto a dispersive element, which decomposes it into its spectrum. The spectrum is then sampled and recorded using either single or multiple detectors. Typically, a diffraction grating is used as the dispersive element because it provides an almost linear relation between wavelength and displacement in the detector plane as opposed to an optical prism, for which the correspondence is highly nonlinear. The linear relationship considerably simplifies calibration procedures.

Modern spectroradiometers use charge-coupled device (CCD) arrays as the detectors because of their linear characteristics. A sampling of the spectrum is achieved automatically through the placement of physically distinct detectors in the measurement plane. Since the separation between the detectors need not correspond directly to a convenient wavelength spacing, the detector outputs are usually interpolated to obtain the final spectral samples. Even though the CCD's are almost linear in their response at a given wavelength, their spectral sensitivity is not uniform. Therefore, a gain compensation procedure is usually necessary to obtain calibrated output from the device [175, p. 338].

The range and the sampling interval of spectroradiometers vary according to their intended application. Spectroradiometers used for color typically report measurements over the range of 360–780 nm and are capable of a spectral resolution of 1 to 2 nm. This resolution is sufficient for most radiant spectra. However, since some light sources have monochromatic emission lines in their spectra, a deconvolution of the spectroradiometer measurements may sometimes be necessary to obtain greater accuracy [27], [28].

An interesting application of spectroradiometry that extends beyond the visible spectrum is in multispectral scanners carried by remote sensing satellites. These scanners disperse radiation into different spectral bands in much the same way as the spectroradiometers discussed above. Early cameras in these satellites used five to 12 spectral bands extending from the visible into the infrared region [176], [177]. The Airborne Visible Infrared Imaging Spectrometer (AVIRIS) [178], which samples the range of 400–2500 nm at 10 nm resolution, is an example of more recent scanners that use a larger number of bands.

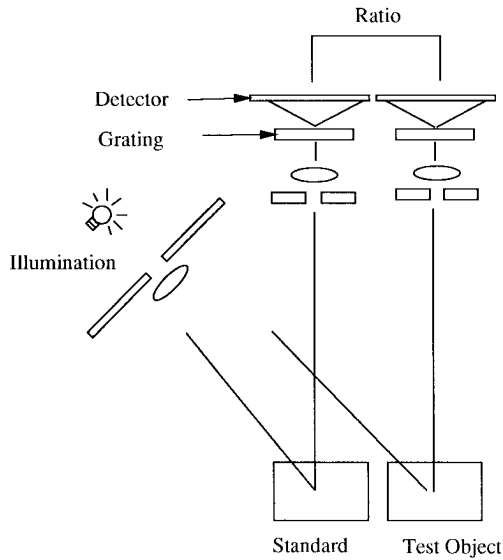

Fig. 10. Spectrophotometer measurement.

The spectroradiometer is useful for measuring the spectra of luminous objects. For nonluminous objects, the spectral reflectance or spectral transmittance is usually more useful for color specification. The device used for measuring spectral reflectance is called a *spectrophotometer*. Since light is essential for making any measurement, spectrophotometers determine spectral reflectance as the ratio of two spectroradiometric measurements. This is shown schematically in Fig 10. The light source is contained within the spectrophotometer and is used to illuminate both a standard sample with known reflectance, $r_s(\lambda)$, and the test object whose reflectance, $r_o(\lambda)$, is to be measured. If $l(\lambda)$ denotes the spectral irradiance of the light source and the device makes $K$ spectral measurements at $\triangle\lambda$ wavelength intervals in the region $[\lambda_0, \lambda_0 + (K - 1)\triangle\lambda]$, the reference measurements can be represented as $m_s(k) = d_k l(\lambda_0 + k\triangle\lambda) r_s(\lambda_0 + k\triangle\lambda)$, $0 \le k \le (K-1)$, where $d_k$ denotes the detector sensitivity at $(\lambda_0 + k\triangle\lambda)$. Similarly, the object measurements are given by $m_o(k) = d_k l(\lambda_0 + k\triangle\lambda) r_o(\lambda_0 + k\triangle\lambda)$, $0 \le k \le (K-1)$. The object reflectance samples can therefore be determined as

$$r_o(\lambda_0 + k\triangle\lambda) = \frac{m_o(k)}{m_s(k)} r_s(\lambda_0 + k\triangle\lambda) \quad 0 \le k \le (K-1).$$

Mathematically, it can be seen that the detector sensitivity, $d_k$, and the illuminant, $l(\lambda)$, cancel out and have no impact on the measurement. However, in order to obtain good performance in the presence of quantization and measurement noise and errors due to the limited dynamic range of the detectors, it is desirable that the product of these quantities be nearly constant as a function of wavelength. For similar reasons, it is desirable that the reflectance of the standard sample be close to unity at all wavelengths. To avoid unnecessary duplication of the optics and sensors, the measurements of the reference standard and the object are usually performed sequentially instead of the parallel scheme shown in Fig. 10. In addition, for added convenience and to save time, typical measurement devices make one measurement of the standard, which is stored and used for a number of successive object measurements.

Since most real-world reflectances are relatively smooth functions of wavelength [27], most spectrophotometers work with much larger sampling intervals than spectroradiometers, typically reporting reflectance at 5, 10, or 20 nm intervals. The built-in illumination in these devices is usually a filtered incandescent or xenon arc lamp whose spectrum is smooth (unlike fluorescent lamps) and therefore does not unduly amplify the measurement noise and quantization errors. Spectrophotometers used in color work usually sample the spectrum in the 380–780 nm range, though the lower wavelength end of the spectrum may be truncated or less accurate in some devices because of the lower energy in incandescent lamps at the ultraviolet end. Owing to the lower resolution requirement and because of the less stringent calibration required (due to the normalization of illuminant and detector sensitivities), spectrophotometers are considerably less expensive than spectroradiometers, and are also more stable over time.

The design of spectroradiometers and spectrophotometers needs to account for a large number of factors excluded from the simplistic description given above. Both devices suffer from systematic and nonsystematic deviations from the ideal behavior described above and need to be accurately calibrated to known radiant and reflectance standards prior to use. In particular, stray light, detector nonlinearity, effects of polarization, variations in illumination and measurement geometry, and unaccounted fluorescence and thermochromism of samples are sources of systematic errors. Detector noise and quantum fluctuations in photon flux are examples of random errors encountered in measurements. The reader is referred to [179, Chap. 9] and [135, Chap. 8] for a thorough, though somewhat dated, account of the systematic errors in these devices and their calibration procedures. A more current, though brief, overview is also provided in [60, Chap. 5] and [141, pp. 74–86]. Detector noise models for older instruments that used thermal detectors and vacuum tubes are described in [180], and a more recent account of noise models for semiconductor detectors of radiation is provided in [175] and [181]–[183]. Some methods for accounting and correcting some of the systematic errors in spectrophotometers are discussed in [184]. The propagation of spectrophotometric errors in colorimetry has also been analyzed in [185].

Color recording devices that attempt to sample spectral information suffer from a number of obvious drawbacks. First, since the total energy in the spectrum is split into a number of spectral samples, a sizeable measurement aperture and/or long integration time are required for reliable measurements of the spectra. In addition, the required optical components make some of the spectral devices rather expensive and therefore inappropriate for desktop use. Finally, measurement devices that exploit trichromacy are less accurate but can offer acceptable color performance and significant speedup at a fraction of the cost. Spectroradiometers and spectrophotometers are therefore used primarily for color calibration, where the larger aperture and longer measurement times are not prohibitive (in contrast with devices for recording complete spatially varying images).

*2) Photographic Film-Based Recording Schemes:*
Photograpihic film is not a digital recording device; however, a brief discussion of this medium is worthwhile, as it often forms the primary input to many digital color imaging systems. Film used for color photography records the color information in three spectral bands corresponding roughly to the red, green, and blue regions of the spectrum.

The image to be recorded is focused by a lens onto the film. The film contains three emulsion layers with silver halide crystals that act as the light sensors and sensitizing dyes that make the crystals in the three layers respond to different spectral regions. Typically, the top layer is blue sensitive, followed by a yellow filter and green- and red-sensitive layers, respectively. The yellow filter keeps blue light from getting to the lower layers that are also sensitive to blue light. Light in each of the three spectral bands initiates the formation of development centers in the corresponding film layer. When the film is chemically processed, the silver halide crystals at the development centers are converted into grains of silver and unexposed crystals are removed. The number of grains of silver in a given layer at a particular location is determined by the incident light energy in the image in the corresponding spectral band at that location. Thus, the spatial distribution of silver grains in the three layers forms a record of the spatial distribution of blue, green, and red energy in the image.

The relation between the density of silver grains and the incident light spectrum is highly nonlinear. In addition, the formation of silver grains is not deterministic, and the randomness in grain formation contributes to noise in the recording process, known as *film grain noise*. Film grain noise is often modeled as a Poisson or Gaussian random process [186, pp. 619–622], [187]–[189] and constitutes multiplicative noise in the recorded image intensity [52, p. 342].

An image record in the form of three layers of silver grains is obviously of limited use. Therefore, further chemical processing of the film is necessary. For the purposes of this discussion, it suffices to note that this processing replaces the silver grains in the red, green, and blue layers with cyan, magenta, and yellow dyes in accordance with the principles of *subtractive color reproduction*, which were be discussed in Section III-A2. A more complete description of color photography can be found in [130], and simplified mathematical models for the process are described in [52, pp. 335–339].

As an aside, one may note that prior to the invention of spectrophotometers and spectroradiometers, two techniques were developed to record the spectral information of entire images on (monochromatic) film. In the microdispersion method of color photography, the light from each small region of image was split into its spectral components using dispersive elements, and the corresponding spectra (of rather small spatial extent) were recorded on film. The second method, known as *Lippman photography*, recorded the color information in the form of a standing wave pattern by using a mercury coating on the rear of the film as a mirror. Both methods required extremely fine-grain film in order to achieve the high resolution required and long exposure times to compensate for the low energy at each spectral wavelength. The reader is referred to [130] for a slightly more detailed account of these techniques.

*3) Colorimeters, Cameras, and Scanners:* Colorimeters, digital color cameras, and color scanners are color recording devices that operate on very similar principles and differ primarily only in their intended use. All these devices record color information by transmitting the image through a number of color filters having different spectral transmittances and sampling the resulting "colored" images using electronic sensors.

The colorimeter is primarily intended for color calibration or quality control applications and is used to measure the color (typically using the CIE system) of luminous (or externally illuminated) objects of somewhat larger angular subtense. Thus, these devices do not involve any spatial sampling, have one sensor per color channel, and make a single average color measurement over their aperture. Colorimeters are often used for the calibration of color monitors.

Digital color cameras are designed to capture color images of real-world objects in much the same way as conventional cameras, with the difference that the images are recorded electronically instead of using film. Since the scenes may involve moving objects, they typically have 2-D CCD arrays that capture the image in a single electronically controlled exposure. Different schemes may be used to achieve the spatial sampling and color filtering operations concurrently. One arrangement uses three CCD arrays with red, green, and blue color filters, respectively. In such an arrangement, precise mechanical and optical alignment is necessary to maintain correspondence between the images from the different channels. Often the green channel is offset by half a pixel in the horizontal direction to increase bandwidth beyond that achievable by individual CCD's [190]. For economy, and in order to avoid the problems of registering multiple images, another common arrangement uses a color filter mosaic that is overlaid on the CCD array during the semiconductor processing steps. Since the green region of the spectrum is perceptually more significant, such mosaics are laid out so as to have green, red, and blue recording pixels in the ratio $2:1:1$ or $3:1:1$ [191]. Image restoration techniques are then used to reconstruct the full images for each of the channels [192]–[194].

Scanners are usually designed for scanning images reproduced on paper or transparencies and include their own sources of illumination. Since the objects are stationary, these devices do not need to capture the entire image in a single exposure. Typical drum or flatbed moving stage scanners use a single sensor per channel, which is scanned across the image to provide spatial sampling. The single sensor makes the characterization of the device easier and more precise, and also allows the use of more expensive and accurate sensors. For desktop scanners, speed is of greater importance, and therefore they usually employ an array of three linear CCD sensors with red, green, and blue color filters. The linear sensors extend across one dimension of the scanned image. This allows three filtered channels of the image along a line to be acquired simultaneously. To sample the entire image, the linear array is moved optically or mechanically across the other dimension of the image. In another variation of these devices, three different lamps are used in conjunction with a single linear CCD array to obtain a three-band image from three successive measurements.

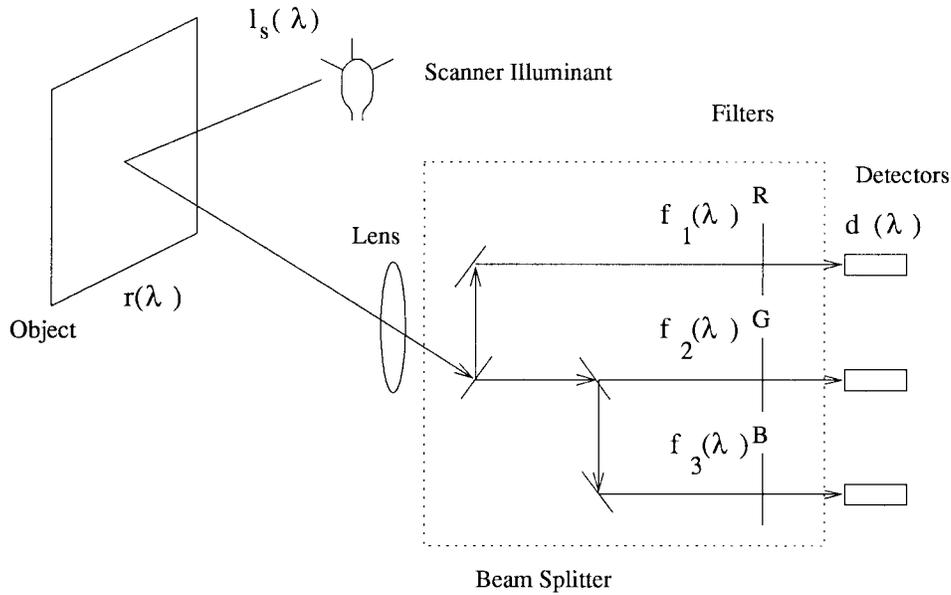

Fig. 11. Schematic of a color scanner.

Colorimeters, digital cameras, and scanners can be mathematically represented by very similar models. In the remainder of this section, a scanner will be used for illustration of such a model. However, the same discussion applies to colorimeters and cameras with trivial modifications that will be pointed out where required.

The schematic of a typical desktop color scanner is shown in Fig. 11. The scanner lamp illuminates the image, and the light reflected off a small area is imaged by the lens onto a beam splitter that splits the light into a number of channels with filters having different spectral transmittances (the typical case of three channels is shown in the figure). The filtered outputs are integrated over the electromagnetic spectrum by optical detectors to obtain a scanner measurement vector. This process is repeated over the entire image to obtain a "color" representation of the image. In actual scanners, the scanner measurements of the small area corresponding to a sampling unit is influenced by the color of the surrounding areas [195]. Ideally, restoration schemes should be used to remove the blur from the recorded image. However, due to the computational requirements, this is rarely done, and this aspect of the problem will be ignored in the subsequent discussion.

For sensors commonly used in electronic scanners, the response at a single spatial location can be modeled in a manner similar to (1) as

$$t_i^s = \int_{-\infty}^{\infty} f_i(\lambda) d(\lambda) r(\lambda) l_s(\lambda) \, d\lambda + \epsilon_i$$
$$= \int_{-\infty}^{\infty} m_i(\lambda) r(\lambda) l(\lambda) \, d\lambda + \epsilon_i \quad i = 1, 2, \cdots K \quad (25)$$

where $K$ is the number of scanner recording channels, $\{f_i(\lambda)\}_{i=1}^{K}$, are the spectral transmittances of the color filters, $d(\lambda)$ is the sensitivity of the detector used in the measurements, $l_s(\lambda)$ is the spectral radiance of the illuminant, $r(\lambda)$ is the spectral reflectance of the area being scanned, $\epsilon_i$ is the measurement noise, $m_i(\lambda) = f_i(\lambda) d(\lambda)$ is the product of filter transmittance and detector sensitivity, and $t_i^s$ denotes the value obtained from the $i$th channel.

In a manner analogous to (2), the equations of (25) may be replaced by their discrete approximations using matrix vector notation as

$$\mathbf{t}_s = \mathbf{M}^T \mathbf{L}_s \mathbf{r} + \boldsymbol{\epsilon} \qquad (26)$$

where $\mathbf{t}_s$ is the $K \times 1$ vector of scanner measurements, $\mathbf{r}$ is the $N \times 1$ vector of reflectance samples, $\mathbf{L}_s$ is an $N \times N$ diagonal matrix with samples of the radiant spectrum of the scanner illuminant along the diagonal, $\mathbf{M}$ is an $N \times K$ matrix whose $i$th column, $\mathbf{m}_i$ is the vector of samples of the product of the $i$th filter transmittance and the detector sensitivity, and $\boldsymbol{\epsilon}$ is the $K \times 1$ measurement noise vector.

Note that while these devices "sample" color spectra very coarsely, to assure that the above model is accurate it is necessary that sampling restrictions on the color spectra involved are met [27]. Due to their higher efficiency and lower heat dissipation, fluorescent lamps are often used in desktop scanners. Since their spectra have sharp spectral peaks, the sampling rate requirements (with uniform sampling) in the model of (26) can be prohibitively high. A more efficient model for such a case is proposed in [196], where a decomposition of the illuminant into the sum of a bandlimited (smooth) component and impulses (monochromatic emission lines) is used to substantially reduce the dimensionality of the model while retaining the mathematical form of (26).

For colorimeters and color cameras, the stimulus is normally a luminous object or an object illuminated by an illuminant external to the device. For these devices, the product, $\mathbf{L}_s \mathbf{r}$ (or its equivalent), defines the spectral radiance whose color is to be recorded. From the model in (26), it can inferred that in the absence of noise, exact CIE XYZ tristimulus values can be obtained from the data recorded by colorimeters and color cameras if there exists a transformation that transforms the sensor response matrix, $\mathbf{M}$, into the matrix of CIE XYZ color matching functions, $\mathbf{A}$ [38]. This is equivalent to the requirement that the HVSS be contained in the *sensor visual space* defined as the column space of $\mathbf{M}$ [26]. For devices using three channels, this reduces to the requirement that $\mathbf{M}$ be a nonsingular linear transformation of $\mathbf{A}$. This fact has been known for some time and is referred to as the *Luther–Ives condition* [197], [198]. Recent reiterations of this result can be found in [199] and [200]. A device that satisfies (generalizations of) the Luther–Ives condition will be said to be *colorimetric*.

For color scanners, the analysis is slightly more involved because the illuminant used in the scanner is usually different from the illuminant under which the scanned object is viewed by an observer. Under these conditions, it can be shown that the CIE XYZ tristimulus values of the scanned object under the viewing illuminant can be determined exactly from the noiseless scanner measurements if the human visual (viewing) illuminant space (HVISS) is contained in the *scanner visual space* (SVS) defined as the column space of $\mathbf{L}_s\mathbf{M}$. Since the spectra of fluorescent lamps used in most scanners is quite different from that of the daylight illuminants used in colorimetry, this condition is rarely met in practice. In addition, often color tristimuli under multiple viewing illuminants need to be estimated from a single scan of the image, and the above criterion would require an inordinately large number of detectors. In addition to the problems caused by fluorescent lamps, actual colorimeters, cameras, and scanners are subject to a wide variety of restrictions arising out of economic considerations and limitations of the processes and materials for manufacturing filters, optical components, and sensors. Techniques from signal processing are therefore useful for the evaluation and design of these devices.

It is the filters, $\{f_i(\lambda)\}_{i=1}^3$, over which the designer has the most control. A quality measure for evaluating single color filters was first proposed by Neugebauer [201]. Recently, this was extended to provide a computationally simple measure of goodness for multiple filters in terms of the principal angles between the HVISS and the SVS [26]. The measure was used for the evaluation and design of color scanning filters [202], [203]. The same measure was also successfully applied to the combinatorial problem of selecting an appropriate set of filters for a scanner from given off-the-shelf candidate filters [204]. A minimum-mean-squared error (MMSE) approach, which requires more statistical information than purely subspace-based approaches, was introduced in [205], where numerical approaches for minimizing errors in uniform color spaces were also considered. In [206], noise was included in the analysis, and [207]–[209] emphasized the reduction of perceived color errors in a hybrid device capable of measuring both reflective and emissive objects through the use of linearized versions of CIELAB space [210]. An alternate novel approach accounting for noise was proposed in [211], where a filter-set was chosen from the multitude satisfying the Luther–Ives condition so as to minimize the perceptual impact of noise. In [212], a unified treatment encompassing a number of these approaches is presented, and their performances are compared.

In actual devices, it is possible to have systematic deviations from the linear model of (26). Sources of error include fluorescence of samples in scanners, stray light, inclusion of ultraviolet and infrared radiation in the measurements (which is not accounted for if the visible region of the spectrum is used in the model), and limited dynamic range of detectors [213]. However, if proper precautions are taken, these errors are small and can be included in the noise process with minimal loss of functionality [213].

It should also be noted here that the above discussion applies to a system for recording color where the input spectra are not constrained to lie in a restricted set. In recording color information from color reproductions that exploit trichromacy and utilize three primaries, the requirements for obtaining precise color information are much less stringent, and typically sensors with any three linearly independent channels suffice. A proof of this result for a system using three additive primaries (whose spectra vary only in amplitude and not in spectral shape) appears in [23]. An example of an application where this can be readily seen is the measurement of colors produced on a CRT [214]. Note, however, that the calibration of these noncolorimetric recording systems is highly dependent on the primaries used in creating the images. Thus, they yield large color errors with images that are not produced with the primaries used in calibration.

For subtractive color reproduction systems (described in Section III-A2) that use varying densities of cyan, magenta, and yellow dyes to reproduce colors, one can conclude that any three sensors from whose measurements the densities can be inferred will suffice. While mathematical characterization of this requirement would require assumptions on the spectra of the dyes and models for the specific processes used, in practical systems any three reasonably narrow color filters with peaks in the red, green, and blue regions suffice [215], [130, p. 247]. Since this is far less demanding as a design objective than the colorimetric criteria discussed above and because a large fraction of input images to scanners are in the form of photographic prints that use subtractive reproduction, most present day scanners are designed to satisfy this requirement. The cost levied by this design trade-off is greater user intervention since distinct calibrations of the scanner are required for accurately scanning reproductions produced with different subtractive primaries [216]. With the advent of hi-fi printing systems that employ more than three primaries, the accuracy of such scanners is likely to be further compromised.

If scanners are designed to be colorimetric, a linear transformation, independent of the scanned object characteristics, can be used to accurately estimate the CIE XYZ tristimulus values from the scanner measurements. However, due to the nonlinear relationship between density and tristimuli, scanners designed to measure dye densities perform poorly with a linear transformation. A number of heuristic nonlinear calibration schemes have therefore been used in practice. Three-dimensional look-up tables [146], least-squares polynomial regression [217], [146], and neural networks [147] are examples of these approaches. Note, however, that these approaches offer significant gains over a simple linear transformation only when the characterization is performed for a restricted class of media [216].

*4) Multispectral Imaging Systems:* In remote sensing applications, multispectral scanners often utilize narrowband spectral filters to record energy in different regions of the spectrum in a manner very similar to the color recording devices mentioned in Section III-B3. A major difference between these and the color recording devices arises from the fact that they are not attempting to capture information so as to match a human observer. Therefore, these devices are not restricted to operating within the visible region of the electromagnetic spectrum and typically use infrared, visible, and microwave regions of the spectrum [177]. For the same reason, while dimensionality reduction of recorded data is often done while processing (see Section III-B5), there is no direct analog of trichromacy in remote sensing.

Most multispectral remote sensing applications are geared toward classification of acquired images into regions belonging to a number of known categories. While this is different from the color recording devices goal of capturing images without loss of visual information, the mathematical model of (26) can also be used to represent multispectral systems. Since the restrictions on the fabrication of these devices are similar to those for color recording devices, a number of ideas mentioned in the previous sections could also be applied to the design and analysis of multispectral imaging systems.

*5) Principal Component Recording:* The color recording devices of Section III-B3 attempt to sample the spectra of images while preserving visual information. A recording of the spectra itself provides greater information but is extremely slow and expensive. Since spectral information of reflective images is extremely useful for determining of color under different illuminants, alternate schemes for recording their spectral information are of interest.

Note that in the absence of noise, the scanned image in (26) can be directly used to determine the projection of the image spectra onto the SVS. Hence, to obtain good reconstruction of reflectance spectra, the sensors can be chosen so that a large fraction of the energy in reflectance spectra lies in the SVS. In the absence of noise, the Karhunen–Loève (KL) transformation provides the mathematical solution to this problem in terms of the statistics of the ensemble of reflectance spectra [218]. The best spectral reconstruction of scanned spectra in a mean-squared error (MSE) sense is obtained from a $K$ channel scanner when the SVS corresponds to the span of the $K$ *principal components* of the reflectance spectra, i.e., the eigenvectors associated with the $K$ largest eigenvalues of the spectral reflectance correlation matrix.

The reflectance spectra of most naturally occurring objects are smooth functions of wavelength; the same is true of spectra produced using photography, printing, or paints. As a result, these spectra can be accurately represented by a few principal components. Various studies of reflectance spectra have estimated that between three to seven principal components (depending on application) provide satisfactory reconstruction of reflectance spectra for most color work [219]–[223]. Note that this offers a significant reduction in dimensionality in comparison with spectrophotometric measurements using uniform sampling.

Linear models for object reflectance spectra based on the principal components idea have been used by a large number of researchers for recovering illuminant and surface reflectance data from recorded images and for color correction applications [220], [223]–[225]. Most of this research used KL transform on a spectrophotometrically recorded ensemble of reflectance spectra, and the problem of designing spectral recording devices based inherently on the principal components approach has received little attention. There is, however, one commercial color measuring device that attempts to reconstruct spectral data from sensor measurements [226]. In addition, the principal components approach has been used in analyzing multispectral satellite imagery, and the idea of a recorder based on principal components has also been suggested for acquiring satellite images [176, Chap. 7].

One may note here that some naturally occurring reflectance spectra do not adhere to the smoothness assumption. Examples of such spectra are colors produced due to multiple film interference in certain minerals and iridescent colors on some bird feathers and in shells containing calcium carbonate [114, pp. 261, 267]. A principal components scheme leads to relatively large errors in such spectra. Hence, in imaging applications involving these objects, the principal components approach would be inappropriate.

## IV. COLOR IMAGE PROCESSING ALGORITHMS

For obvious historical reasons, a large fraction of the existing research in digital image processing deals only with monochrome images. While some of this can be extended to the processing of color images in a straightforward fashion, there are several problems that are unique to the processing of color images. This section is a survey of the research addressing some of these problems.

### A. Color Quantization

Most computer color displays are based on a frame buffer architecture [227]. In such an architecture, the image is stored in a video memory from which controllers constantly refresh the display screen. The images are usually first recorded as *full color* images, where the color of each picture element (*pixel*) is represented by (gamma-corrected) tristimuli with respect to the display's primaries and quantized to 8 or 12 bits (b) for each of the three channels. Often, the cost of high-speed video memory needed to support storage of these full color images on a high-resolution display is not justified. Many color-display devices therefore reduce memory requirements by restricting the number of colors that can be displayed simultaneously. Often, 8, 12, or 16 b of video memory are allocated to each pixel, allowing simultaneous display of $2^8, 2^{12}$, or $2^{16}$ colors, respectively. The user then has the capability to choose a palette of simultaneously displayable colors from a much larger set of colors that the device is capable of rendering. A *palettized image*, which has only the colors contained in the palette, can be stored in the video memory and rapidly displayed using look-up tables implemented in hardware [227].

The use of a fixed image-independent palette usually produces unacceptable results, unless halftoning (see Section

IV-B) is employed. Hence, most image displays use an image dependent palette. In the ideal case, the palette and the palettized image should be simultaneously determined from a full-color image so as to minimize the perceived difference between the displayed and the full-color image. Since this solution is intractable, usually it is simplified by splitting it into two successive steps: i) the selection of a palette and ii) the mapping of each pixel to a color in the palette.

The problem of selecting a palette is a specific instance of the more general problem of vector quantization (VQ) [228], [229]. If the true color image has $N$ distinct colors and the palette is to have $K$ entries, the palette selection may be viewed as the process of dividing $N$ colors into $K$ clusters in 3-D color space and selecting a representative color for each cluster. Ideally, this clustering should minimize perceived color error. For mathematical tractability, however, the problem is often formulated in terms of minimization of the MSE between the true-color tristimuli in the image and the palette representatives of their clusters. The selection of a globally optimal palette under the MSE criterion is a nondeterministic polynomial-time (NP) complete problem [230]–[232]. For cases where the number of palette and true colors is extremely small, a branch and bound algorithm can be used to determine the optimal solution [233]. However, for most realistic images, this approach is infeasible. Therefore, efficient palette selection algorithms use suboptimal formulations of the problem and heuristics to achieve acceptable performance in reasonable time. Often, the Linde–Buzo–Gray (LBG) algorithm [234] can be used to iteratively improve an existing palette to achieve a local minimum with respect to the MSE criterion. The algorithm is a generalization of the 1-D Lloyd–Max quantization algorithm [235], [236] and is identical to the $K$-means clustering algorithm [237] used in pattern recognition and classification.

A simple heuristic that has been used for palette design is the *popularity algorithm*, which works by forming a 3-D histogram of the true image colors and assigning the $K$ most frequently occurring colors in the histogram as the palette colors [238]. While the popularity algorithm is extremely fast, it performs rather poorly on images with a wide range of colors. The idea of using histograms is, however, useful, and is often used as a first step in the palette selection process to reduce the number of colors to more manageable levels. Typically, the histogram is formed by the simple process of ignoring some of the least-significant bits in each tristimulus. Braudaway [239] suggested a variant of the popularity algorithm that prevents the concentration of too many colors around a single histogram peak by allocating the palette colors sequentially and modifying the histogram after each allocation. Gentile *et al.* [124] modified Braudaway's algorithm and used it to perform palettization in CIELUV space. An alternate heuristic was suggested by Heckbert [238], which attempts to use each palette color to represent an equal number of true colors. The proposed algorithm determines a palette by a recursive process of splitting the largest cluster into two equal halves. Since the splitting is done about the median point after sorting the colors in the cluster along the dimension with largest spread, the algorithm is commonly referred to as the

*median cut algorithm*. Both Heckbert and Braudaway used the LBG algorithm to improve the initial palettes obtained from their heuristic procedures. However, since the LBG algorithm converges only to a local minimum, this step often yields only slight improvements [238].

More significant improvements can be obtained by replacing the heuristics by schemes that attempt to select the palette through a sequential splitting process while reducing the MSE at each step. Various splitting procedures, and different criteria for selection of the cluster to be split, have been used by researchers. An algorithm suggested by Wan *et al.* [240], [241] (binarily) splits the cluster with the largest MSE at each stage along the plane orthogonal to the (tristimulus) coordinate axis yielding greatest reduction in the total MSE. A similar scheme has also been reported more recently in [242]. Orchard and Bouman [243] developed a more generalized binary splitting algorithm that allows arbitrary orientations of the plane used to split a cluster. The algorithm selects the cluster with the most variation along a single direction and splits it along the direction of maximum variation. It also incorporated a modification of the MSE by a subjective weighting factor to reduce undesirable visible artifacts of quantization. Balasubramanian *et al.* [244] used efficient data-structures, histogramming, and prequantization to speed up the Orchard–Bouman splitting algorithm.

Since the binary splitting algorithms use the "greedy" strategy of minimizing the MSE at each split, they can potentially get stuck in poor local minima. Wu [245] developed an alternative dynamic programming scheme that performs multiple splits at each stage to partially remedy this problem. For reasonable sized color palettes, the potential for encountering poor local minima with the greedy strategy is rather low. Consequently, Wu's algorithm offered the greatest gains for quantization with small palettes [245].

While several of the VQ algorithms mentioned above allow nearly transparent quantization of images for display, their computational cost is often too high. Recently, Balasubramanian *et al.* [246] reported a new VQ technique called sequential scalar quantization (SSQ). SSQ is able to exploit correlation between the color components and offers some of the benefits of conventional VQ, while retaining the simplicity of scalar quantization. The application of SSQ to color quantization has been reported in [246] and [247], where it can be seen that SSQ offers slightly inferior MSE performance in comparison to some of the binary splitting techniques while providing very significant speed-up. SSQ has also been used for the creation of universal (image-independent) color palettes for error-diffusion (discussed in Section IV-B) [248].

One limitation of the palettization schemes based on MSE is that they offer no guarantees regarding the maximum color error. The octree quantization algorithm [249], [250] and the center-cut algorithm [251] are two simple color quantization schemes motivated by the idea of limiting the maximum error. The center-cut algorithm is a minor modification of the median-cut algorithm, in which the cluster with the largest dimension along a coordinate axis is split along the center at each step. The octree algorithm is a bottom-up approach to the problem, in contrast with all the other top-down schemes.

Conceptually, it subdivides the color space into cubes until each cube contains only one image color and then reduces the resulting octree by an averaging and merging process so that each node represents a palette color. In practice, the building and reduction are performed in a single pass. An alternate bottom-up color quantization scheme, that uses histogramming and prequantization with a cluster-merging VQ algorithm, was also presented in [252].

Thus far, the palettization algorithms discussed were concerned with single images. The problem of palettization is more involved for video sequences. In addition to the requirement of real-time performance, care must be taken in the design of palettes for successive frames to minimize the visible effects of *colormap flashing* that occurs when the color map is updated prior to the update of a frame. The reader is referred to [253] for a description of a mathematical formulation and solution scheme for the problem of color quantization of video sequences. A less sophisticated scheme utilizing a single color palette for the entire sequence is also discussed in [254]. Colormap flashing can also be encountered in the simultaneous display of multiple independently palettized images. The problem can be eliminated through the use of a combined palette. An efficient VQ scheme for combining palettes with little visual distortion is presented in [255].

Once the design of a palette is over, the second step in color quantization, i.e., the mapping of image pixels to the palette colors, needs to be performed. The simplest approach to this problem is to map each pixel to its nearest neighbor in the palette. Often, the palette design process can be used to obtain a tree structure that simplifies this nearest neighbor search or provides a good approximation to it. This is particularly true for the binary splitting algorithms that can use the classic k-d tree [256] for nearest-neighbor search. Similar procedures can also be readily used with the octree quantization scheme. Note however, that if the LBG algorithm is used for iterative improvement of the obtained palette, the tree structure of the palette is destroyed, and therefore the nearest-neighbor search is more involved though fast searching schemes can still be developed [257], [258].

While the nearest-neighbor mapping algorithm is optimal from a minimum-average-error standpoint, it often produces objectionable contours in smooth image regions. A significant reduction in visible contours can be obtained by a using the halftoning techniques of dither or error diffusion, which will be briefly discussed in Section IV-B. The use of these techniques in the pixel mapping step has been discussed by a number of researchers [238], [239], [124], [243]. Liu *et al.* [242] describe another pixel mapping scheme that combines error diffusion with morphological operations in an attempt to reduce visible artifacts.

### B. Halftoning

The human eye is extremely sensitive to color variations, and is capable of distinguishing around 10 million colors under optimal viewing conditions [259]. At the same time, color output devices such as halftone color printers and palette-based displays are capable of producing only a limited number of colors at each addressable spatial location. However, these devices normally possess rather high spatial resolution, which (at normal viewing distances) is often beyond the resolving capabilities of the human eye. In reproducing color images on these devices, it is therefore desirable to use techniques that trade off (excess) spatial resolution in favor of a greater range of perceived colors. The term *halftoning* is used to describe a variety of image processing techniques based on this idea.

The eye perceives only a spatial average of the microvariation in spot-color produced by the device, and is relatively insensitive to high-frequency differences between the original continuous-tone image and the halftone image. Halftoning algorithms therefore attempt to preserve this average in the reproduction while forcing a large fraction of the (necessary) difference between the halftone and contone images into the perceptually irrelevant high-frequency regions. In the remainder of this section, the distinction between spot-color and perceived average color will not always be explicitly emphasized. The implied meaning should, however, be clear from the context. Similarly, most of the discussion will refer to halftone printing, and the use of halftoning in displays will be mentioned where appropriate.

Halftoning methods have been used in lithographic printing for the reproduction of both gray-scale and color images for a considerable length of time [260, p. 128]. The halftones for lithographic printing processes were traditionally obtained by photographing (color-filtered) images through a fine screen on a high-contrast film [260, Chap. 7]. In digital imaging applications, halftoning was originally used in binary display devices and printers for producing the illusion of gray scale. There is a vast amount of literature dealing with the halftoning of gray-scale images (see [261]–[265] for an extensive bibliography). This section will focus mainly on digital color halftoning. Since several techniques of color halftoning inherit their motivation and principles from prior schemes used in halftoning gray-scale images, some of the pertinent gray-scale schemes will also be referenced.

One problem unique to color halftoning is the problem of registration. Most printing processes print halftone "separations" of the different color dyes sequentially. Typically, the reproduction medium is moved by mechanical systems in the process and some variation in the alignment of these separations is inevitable. If ideal cyan, magenta, and yellow dyes (which follow the Bouguer–Beer law and the block dye assumptions mentioned in Section III-A2) are used, the visual appearance would be essentially independent of small registration errors [134]. However, real dyes are far from ideal, and registration errors can produce significant color shifts due to varying amounts of overlap between the separations. Color halftoning schemes, therefore, attempt to arrange separations so that the relative overlaps of the dye layers are insensitive to alignment errors.

In conventional digital color halftoning (for printers), the image is decomposed into a cyan, magenta, yellow, and black separations, which are halftoned independently. The halftoning for each separation is done by comparing the pixel value with a spatially repeated *dither* array and turning on pixels for which the image exceeds the value in the corresponding dither matrix [264]. The dither matrix is designed so that thresholds close in value occur near each other in the dither array. For a uniform image, the halftoning schemes yield a grid of halftone dots (consisting of clusters of "on" pixels) similar to one produced by the photographic screen in lithography. The size of the *clustered* dots increases with increase in the image pixel values. The grids for different colors are oriented at different angles to make the overlaps between the dots in the four separations invariant to small registration errors. The printing mechanism is therefore said to employ *rotated screens*. The rotation angles for the different colors are chosen so as to minimize the occurrence and visibility of low-frequency interference patterns, known as *moiré* [134]. The most visible black screen is typically oriented along a 45° angle, along which the eye is least sensitive. The yellow, magenta, and cyan screens are located along 0°, 15°, and 75°, respectively [134, pp. 328–330]. In digital imaging applications, often one is confined to working on a rectangular raster. An elegant scheme for obtaining different screen angles on rectangular rasters was developed by Holladay [266]. An analysis of moiré using Fourier transforms and methods of designing dither arrays that minimize moiré can be found in [267] and [268].

The requirement that pixels in a dither pattern must be clustered together is fairly restrictive and compromises spatial resolution. However, clustered dots are insensitive to common printing distortions and reproduce well on printers and copiers that have difficulty in reproducing isolated pixels [265]. Therefore, rotated clustered-dot screens have been used extensively for color printing. For displays, these considerations are inapplicable, and therefore alternate dither matrices that produce *dispersed* dots with greater spatial resolution have been successfully used for bilevel displays [269]. The use of rotated dispersed-dot dithering for color printing on inkjet printers has also been recently mentioned in [270].

One may note here that the process of thresholding with a dither array can be replaced by a mathematically equivalent scheme of adding a dither pattern to the image and thresholding at a constant level. This is a variant of the scheme proposed in [271]. The original scheme proposed the use of random noise as the dither pattern. Such a scheme is known to reduce visible artifacts due to quantization and is often used in monochrome and color displays. In order to emphasize the difference with this random dither, the term *ordered dither* is often used to describe schemes of the last two paragraphs, for which the dither pattern is not random.

Halftoning schemes that employ dither matrices quantize the image pixels in isolation. Considerable improvements in image quality can be obtained by using adaptive schemes that process each pixel depending on the result of processing other pixels. Error-diffusion (ED) [272], [273] is an adaptive scheme that has been widely used. ED works by "diffusing" the error resulting from the quantization of the current pixel to neighboring pixels. At each pixel, the diffused error is added to the image value prior to quantization, and the quantization error is again distributed over adjacent pixels. Since the objective of error diffusion is to preserve the average value of the image over local regions, a unity-gain lowpass finite impulse response (FIR) filter is used for distributing the error. From a signal processing viewpoint, ED can be viewed as a 2-D version of $\Sigma$-$\triangle$ A/D converters. An analysis of ED and $\Sigma$-$\triangle$ A/D conversion in a common framework was presented in [274], [275]. From such an analysis, it can be seen that the image resulting from ED can be represented as the sum of the original image and a highpass filtered error image [276], [277]. Since the eye is less sensitive to high spatial frequencies, the images resulting from ED typically appear closer to the originals than those obtained with ordered dither.

ED was originally used and analyzed for gray-scale images. Most of the analysis is, however, applicable to color images too. For color images, the image may be represented as tristimuli in a color space or as separate cyan, magenta, yellow, and black images for printing. Typically, the error diffusion is done independently for each channel using identical spatial filters, but quantization may be performed independently in each channel (scalar ED) or simultaneously for the entire color vector (vector ED). For printers, both scalar and vector ED schemes have been used [278], [279], but for color displays with an image-dependent palette, it is usually necessary to use vector ED. For unity-gain ED filters with positive weights only, it can be shown that scalar ED is a stable process with bounded quantizer overload [274]. However, for vector ED, severe quantizer overload can occur, leading to significant color artifacts. A discussion of this problem and methods that attempt to reduce overload by reducing the feedback in ED can be found in [243] for displays and in [279] for printers.

The ED filter needs to be causal if the image is to be halftoned in a single pass. The causality requirement for normal raster processing implies that the filter is asymmetric. This asymmetry often results in visible low-frequency "wormlike" artifacts in bilevel ED. Several schemes, such as a larger area of support for the FIR filter [280], [281], processing on a serpentine raster [282], and symmetric error diffusion neural networks [274], have been proposed for gray-scale images to overcome these limitations. For color images, however, the problem of visible artifacts is not that acute due to the multiple output choices for each pixel in both displays and printers [283]. While clustered dot dithering schemes are often preferred for printing for reasons mentioned earlier, ED is the primary halftoning scheme used in displays.

It was mentioned earlier in this section that ED images are pleasing to the eye because of the highpass nature of the "noise" in the reproduction. The fact that image quantization noise to high spatial frequencies results in reduced noise visibility was recognized early on and exploited in dithering schemes by several researchers [269], [284]–[286]. Noise processes having only high spatial frequency components were given the name *blue noise* by Ulichney [287]. He convincingly argued that since a large MSE is inevitable when reproducing a gray-scale image on a bilevel device, halftoning should attempt to concentrate on shaping resulting noise spectrum to be blue, and therefore least visible. While ED produces blue noise, it offers only a limited control over the noise spectrum and also requires considerable processing in comparison to the pixelwise thresholding for ordered dither. Several researchers have therefore worked on developing large dither matrices, which achieve ED-like performance with the computational benefits of pointwise processing [288]–[290].

The resulting techniques along with ED-like adaptive algorithms are collectively known as *blue noise halftoning* or *stochastic screening*. The dither arrays produced for obtaining blue noise characteristics are known as *blue noise masks*. Most of the literature dealing with the design of blue noise masks addresses only grayscale halftoning. For color, independent (uncorrelated) blue noise masks are used for each separation. Just as in bilevel reproduction, blue noise masks in color offer the advantages of high processing speed, good spatial resolution, and few visible artifacts. An added advantage, particularly for hi-fi printing, is the lack of visible screen texture and interference between angled screens.

The halftoning methods mentioned thus far are (intelligent) heuristic algorithms that exploit the lowpass nature of the eye's spatial response. The problem of halftoning can alternately be formulated as an optimization problem that aims at maximizing the "visual similarity" between the original image and the halftoned image. In order to quantify visual similarity, several schemes utilize a model for the spatial response of the eye. These schemes are therefore referred to as *model-based halftoning methods*. Based on psychophysical measurements with sinusoidal gratings, an empirical isotropic linear shift-invariant (LSI) model for the eye's spatial response was developed in [291] along with a distortion measure. The distortion measure corresponds to the MSE between the filtered versions of the actual and halftoned images, where the filter is the LSI model of the eye's spatial response. Most model-based halftoning methods use variants of this distortion measure. For the spatial response of the eye, however, a number of other models have also been proposed, including some that take into account known anisotropy in the visual-system [292].

The exact optimization in model-based halftoning schemes is an intractable integer programming problem. For display of gray-scale images on binary output devices, a number of researchers have suggested iterative schemes that provide good solutions with varying computational requirements [292]–[296]. For color displays with reasonable sized palettes, the problem remains computationally infeasible at present. An iterative algorithm for model-based halftoning for color printers has been reported in [297]. The algorithm uses a linearized version of CIELAB space for perceptual uniformity along with separate spatial models for luminance and chrominance channels. Pappas [283], [298], [299] has also considered an extension of the model-based halftoning scheme to color that also accounts for some printing distortions.

Since the computational requirements of most model-based schemes are rather restrictive, a number of researchers have investigated hybrid schemes that use the halftoning methods mentioned earlier but still attempt to minimize a visual model-based error. For gray-scale images, Sullivan *et al.* [300] have incorporated a visual model in ED. The use of neural nets to minimize a visual model-based distortion function in symmetric error diffusion has also been reported in [274]. For color applications, significantly less research has been done. The optimization of ED for color display applications has been reported in [301] and [302], where optimal ED filter coefficients were determined through a process of autoregressive (AR) modeling of the eye's spatial response.

The method of ED continues to be an active area of research. Recent novel developments in the area include the use of adaptive signal processing techniques and embedded quantization schemes. Image-adaptive ED filters that use the well-known least-mean-squares (LMS) algorithm [303] have been employed for halftoning gray-scale [304] and color [305], [306] images (although, with slightly different error criteria). The methods exploit the local characteristics of the image to obtain improvements over a constant filter. Another modification of color vector ED is proposed in [307], [308]. The resulting embedded multilevel ED algorithm allows coarse quantizations of an image to be embedded in finer quantizations, which can be useful in (among other things) the progressive transmission of color images.

One may note here that the issue of obtaining colorimetrically accurate reproduction has been consciously avoided in the discussion above. Due to gamut restrictions, a colorimetric match is often neither feasible nor desirable. In addition, even with these problems excluded, there have been only a few attempts at incorporating colorimetric matching into halftoning algorithms. If the output device is linear and a mean preserving scheme such as conventional ED is used for halftoning in the device color space, it can be expected that a good colorimetric match will be obtained. However, these assumptions are rarely valid. For CRT displays, due to the larger number of output quantization levels, an argument of local linearity can be invoked [309]. This justifies and explains the relatively accurate color reproductions obtained using ED on CRT displays. In printing, color accuracy is normally addressed (using the methods mentioned in Section III-A3) after selecting a halftoning scheme. The problem of colorimetric match can also be partly addressed using a uniform color space (UCS) for vector ED [297], [279]. However, since the number of possible output pixel colors for printers is small, analysis of colorimetric behavior of halftoning algorithms needs further research.

### C. Color Image Coding

Since color has become such a large part of digital imagery, the problem of coding color images for transmission and storage has gained increased importance. The natural evolution of coding treated color images simply as three independent monochrome bands. This allowed all known monochrome coding methods to be used. Since CRT displays were the primary intended target, the RGB representation was most common. However, the monitor-based RGB tristimuli are highly correlated and therefore not suited for independent coding [310]. This was also recognized in early work with color television, and other color spaces were investigated. A review of this work is given in [42]. There are a few important points about this early work that are worth noting here. The importance of using a luminance chrominance space similar to CIELAB and CIELUV was recognized. An independent luminance channel was also required for reasons of compatibility with existing monochrome system. The result was YIQ, which is still used for many applications outside of television today. The YIQ signals had to be transformed into driving signals for the RGB phosphor guns used in the receiver. Since precision analog hardware was expensive, an attempt was made to minimize the complexity of the various transformations. The spectral sensitivities of the recording cameras were chosen to approximate the color-matching functions defined by phosphor primaries. In doing so, the negative portions of the color-matching functions were ignored, and sensitivities that matched the nonnegative portions were used [42]. This was recognized as creating color errors; however, the errors seemed tolerable. In order to simplify receivers, the YIQ signals were derived from gamma-corrected RGB signals associated with the phosphor primaries, and therefore the YIQ space used in TV transmission is not a linear transformation of the CIE XYZ space.

With the advent of HDTV, the problem of negative lobe truncation in the camera sensitivities is to be entirely eliminated by implementing matrix transformations of recorded color tristimuli. The gamma-correction mentioned in Section III-A1 has still been retained due to its perceptual benefits. Instead of YIQ, an opponent color space encoding for gamma-corrected RGB data has been standardized as the YCrCb space [51]. This space has also been used frequently in recent work on image compression.

Since it was well known that color perception errors did not correlate well with Euclidean distances in RGB space and the television standard was available, most research on coding used variations of monochrome coding in combination with transformations to better color spaces. Transformations to novel color spaces were used, such as the K–L transform, which offered performance similar to YIQ [310]. Some work has also been done with coding images using perceptual error measures (see for e.g., [311], [312]). While the CIELUV and CIELAB spaces were designed to match the results of perceptual tests on larger patches of color, tests confirm that quantizing in these spaces produces smaller perceptual errors in images.

With the increasing use of digital imagery that is independent of TV, new methods are continually being introduced. Just as in halftoning, coding schemes that exploit the lowpass spatial response of the eye have been suggested for image coding. In subband coding [313], the image is split into orthogonal subbands with varying frequency content. The subbands are then quantized with fewer bits allocated to higher frequency components. VQ techniques, similar to those described in the earlier section on palettization, can be used for the quantization. Known anisotropy in the eye's spatial response can also be utilized. The lower sensitivity of the eye along the 45° angle, permits fewer bits to be allocated to frequency bands located on the diagonal. This idea is easily extended to color images. However, complete data on the spatial frequency response of the eye to spatial color (chromatic) frequencies has only recently been published. The combination of subband coding with color spatial frequency response was presented in [314] and [315].

Similar ideas provide the motivation for the discrete-cosine-transform-based (DCT-based) Joint Photographers Expert Group (JPEG) and Moving Pictures Expert Group (MPEG) standards [316]–[318]. These standards specify the coding to be performed on each image band, and allow for a variety of color spaces to be used. This means that as better color-coding methods are developed, they can be implemented in the current framework. In current implementations, the YCrCb space is often used with the Cr and Cb components subsampled by a factor of two along both spatial dimensions [319]. The JPEG compression scheme has also been incorporated in the International Color Facsimile Standard [320], [321].

An interesting problem is the coding of palettized images. While the palettization process offers some compression (typically 3 : 1), this is usually significantly lower than what is attainable with other image coding schemes. Smoothness assumptions are typically invalid for color mapped "image" data. Therefore, normal coding schemes are inapplicable unless the images are remapped to full color images before coding. Recently, Wu [322] has suggested a new YIQ palette architecture that uses joint VQ of spatial and color information to obtain modest compression ratios. A more aggressive coding scheme for palettized images has been suggested in [323], where the colormap data is locally reorganized to obtain smooth blocks, and DCT coding is then utilized. Lossless entropy coding schemes have also been presented recently in [324] and [325].

### D. Gamut Mapping

It was mentioned in Section III-A that color output devices are capable of producing only a limited range of colors defined as their gamut. Often, an image contains colors beyond the gamut of the target output device. In such a case, before the image can be reproduced, it is necessary to transform the image colors to lie within the gamut. This process is referred to as *gamut mapping*. The goal in gamut mapping is to obtain a reproduction that appears identical to an "original" image. The original image is often itself a reproduction from another color output device. Since output devices have widely varying physical characteristics, they can have significantly different color gamuts. In addition, devices rely on different mechanisms (e.g., emission, transmission, or reflection) to produce color and therefore imply different viewing conditions and consequently different states of adaptation for the observer's visual system. Gamut mapping is therefore a difficult problem in which the issue of device capabilities is interwoven with the psychophysics of color vision.

While the gamut mapping problem has been successfully addressed in the printing and graphic arts industries for considerable time, it has become an active area of research in digital imaging applications only recently. In graphic arts and printing, skilled human operators rely on experience to perform gamut mapping for each image independently. In digital imaging applications, on the other hand, it is desirable to automate as much of the process as possible and make it transparent to the end user.

Due to the large differences in dynamic range of different color devices and due to the normalizing adaptation in the eye, little success can be achieved by gamut mapping schemes that attempt to match tristimulus values. Use of uniform color spaces that incorporate some white point scaling in the specification of colors mitigates the problem of normalization to a limited extent. However, naive schemes that map out-of-gamut colors to the nearest color in a uniform color space or scale the entire image colors to lie within the gamut are also unsatisfactory in most cases. A robust and universal gamut mapping strategy remains an elusive goal. However, several researchers have reported encouraging results from experiments with different gamut mapping strategies. The more successful approaches tend to use uniform color spaces or color appearance models and manipulate color data using perceptual attributes of lightness, hue, and chroma in an attempt to preserve the more important attributes.

Stone *et al.* [158] laid down some principles of gamut mapping that were culled from psychophysics and accepted procedures in graphic arts. For printing images displayed on CRT monitors, they described an interactive gamut mapping strategy involving translation, scaling, and rotation of colors in CIE XYZ space. For an identical scenario, simulations of a number of clipping and compression based gamut mapping schemes using CIELUV [326], [327] and CIELAB [328], [329] color spaces have also been reported. For obtaining similar reproductions on transmissive and reflective media, the use of an invertible color appearance model was reported in [330]. Recently, a gamut mapping strategy for printers that does not involve any explicit clipping and scaling was presented in [331]. The mapping for a limited subset of colors was explicitly specified, and an interpolation algorithm based on morphing was then used to obtain a mapping for the other colors.

### E. Device-Independent Color and Color Management Systems

In digital imaging applications, color was first used primarily on CRT displays. The color data in most images was adjusted and stored in a device-dependent format suited for providing reasonable reproduction on common CRT monitors. With the proliferation of a large number of other color devices, it is desirable to use a device-independent color specification from which identical-appearing images can be created on multiple output devices. While this has been an active area of research in the industry, a universally acceptable color specification system that guarantees device independence is yet to be defined. The use of standardized colorimetry defines colors in a device-independent space and therefore forms a first step toward achieving device independent color. However, as discussed in Sections II-D and IV-D, the use of a device-independent color space does not by itself guarantee an appearance match between images reproduced on different devices under different viewing conditions. Therefore, in addition to colorimetry, the use of auxillary information regarding viewing conditions and white points has been proposed for color specification. Such information would allow exploitation of vision psychophysics and gamut mapping to achieve device-independent color [332], [333].

For device-independent color reproduction, it is necessary to accurately characterize each individual color input and output device and transform image data into appropriate device-dependent versions based on the characterization. In order to isolate the end user from the nitty-gritty of handling color characterization information and transformations, color management systems have been proposed to automate this task. For a discussion of the systems issues in color management and existing color management schemes, the reader is referred to recent presentations on the topic [334]–[336]. A notable advance in this direction is the emergence of a widely accepted standard [86], [337] to facilitate the communication of device characterizations.

## V. RESEARCH DIRECTIONS IN COLOR IMAGING

This paper surveyed the current technology and research in the area of color imaging. As compared to monochrome imaging, the field of color imaging is still in its infancy and abounds with a large number of interesting research problems. Of greatest relevance, perhaps, is the problem of colorimetric recording of image data. Since the "garbage-in garbage-out" paradigm still holds, significant gains in processing and display can be made only if the recorded colors are accurate. Toward this end, the construction of readily manufacturable and inexpensive colorimetric filters remains elusive, though advances in electrically tunable acouto-optic filters [338], [339, Chap. 7] offer considerable promise. Using appropriate electrical modulation, a wide range of filter transmittances can be synthesized on these devices [340]. Their use in scientific applications requiring precise color recording has been recently reported in [341]. Other research areas in color input systems are the development of multidimensional image-restoration schemes for effective deconvolution of adjacency effects in scanning, robust methods for scanning halftone images, and methods for illuminant independent recording of reflective and transmissive images. Research in several of these areas would also be relevant to satellite multispectral imagers. In quantifying color recording accuracy, it is also desirable to develop metrics that are based on complex image scenes instead of the CIE metrics based on large uniform areas.

Color appearance modeling for imaging applications is another prospective area for investigation. While there are several color appearance models, most are fairly complex and their suitability for color imaging remains to be comprehensively evaluated. The development of simpler and readily applicable (though, not necessarily physiological) models of color perception and their incorporation into color processing algorithms to improve performance is also a desirable research goal. This is particularly relevant for problems of gamut mapping and cross-device rendering, though such an approach would also benefit coding and compression algorithms.

With the advent of new display and printer technologies, their modeling and easy calibration will also pose new challenges. Few predictions, if any, can be made regarding the nature of devices yet to come. However, it is likely that a number of these will utilize more than three primaries to obtain an increased gamut. Color coordinates, which drive the primaries in these systems, would therefore be mathematically underdetermined. Research is needed on schemes for dealing with this underdeterminacy that incorporate feasibility constraints and also avoid introducing undesirable discontinuities, while exploiting the full device gamut. Several of these problems are already under investigation in the context of hi-fi printing mentioned in Section III-A4.

Finally, there is room for improvement in existing color processing algorithms. More efficient quantization and halftoning schemes are necessary for use with real-time video on frame-buffer displays. Colorimetric behavior of halftone printing also deserves attention, though such an analysis would necessarily have to account for imperfections in the printing process in order to be useful.

## ACKNOWLEDGMENT

The authors thank Prof. J. P. Allebach, Dr. P. W. Wong, Prof. M. D. Fairchild, and Prof. B. V. Funt for their helpful comments and suggestions. We would also like to acknowledge assistance received from Dr. R. R. Buckley, Dr. P. L. Vora, Dr. R. Balasubramanian, and numerous other professionals who reviewed early drafts of this paper. Their inputs have helped improve the content and presentation of this paper significantly.

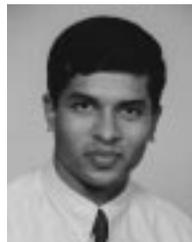

**Gaurav Sharma** (S'88–M'97) was born in Dehradun, India, on October 12, 1968. He received the B.E. degree in electronics and communication engineering from University of Roorkee, India, in 1990, the M.E. degree in electrical communication engineering from the Indian Institute of Science, Bangalore, India, in 1992, and the M.S. degree in applied mathematics and Ph.D. degree in electrical engineering from North Carolina State University (NCSU), Raleigh, in 1995 and 1996, respectively.

From August 1992 through August 1996 he was a Research Assistant at the Center for Advanced Computing and Communications, Electrical and Computer Engineering Department, NCSU. Since August 1996, he has been employed as a member of the Research and Technical Staff at the Digital Imaging Technology Center, Xerox Corporation, Webster, NY. His current research interests include signal restoration, color science, image halftoning, and error correction coding.

Dr. Sharma is a member of Phi Kappa Phi, an associate member of Sigma Xi, and a member of the IEEE Signal Processing Society, IS&T, and SPIE.

**H. Joel Trussell** (S'75–M'76–SM'91–F'94), for a photograph and biography, see this issue, p. 899.